\documentclass[10pt,twocolumn,letterpaper]{article}

\usepackage{cvpr}              

\usepackage{booktabs}
\usepackage{tabularx}
\usepackage{array}
\usepackage{multicol}
\usepackage{multirow}

\usepackage{dblfloatfix}
\usepackage{caption}
\newcommand{\model}{Fast3R}

\DeclareSymbolFont{extraup}{U}{zavm}{m}{n}
\DeclareMathSymbol{\varheart}{\mathalpha}{extraup}{86}
\DeclareMathSymbol{\vardiamond}{\mathalpha}{extraup}{87}

\definecolor{cvprblue}{rgb}{0.21,0.49,0.74}
\usepackage[pagebackref,breaklinks,colorlinks,allcolors=cvprblue]{hyperref}

\def\paperID{3821} 
\def\confName{CVPR}
\def\confYear{2025}

\title{Fast3R: Towards 3D Reconstruction of 1000+ Images in One Forward Pass}

\author{Jianing Yang$^\clubsuit$$^\vardiamond$ \quad Alexander Sax$^\clubsuit$ \quad Kevin J. Liang$^\clubsuit$ \quad Mikael Henaff$^\clubsuit$ \quad Hao Tang$^\clubsuit$ \\
Ang Cao$^\clubsuit$$^\vardiamond$ \quad Joyce Chai$^\vardiamond$ \quad Franziska Meier$^\clubsuit$ \quad Matt Feiszli$^\clubsuit$ \\ \\
$^\clubsuit$ Meta FAIR \quad $^\vardiamond$ University of Michigan \\
\href{https://fast3r-3d.github.io}{\texttt{https://fast3r-3d.github.io}}
\vspace{-20pt}
}

\begin{document}

\twocolumn[{%
    \renewcommand\twocolumn[1][]{#1}%
    \maketitle
    \centering
    \includegraphics[width=0.95\linewidth]{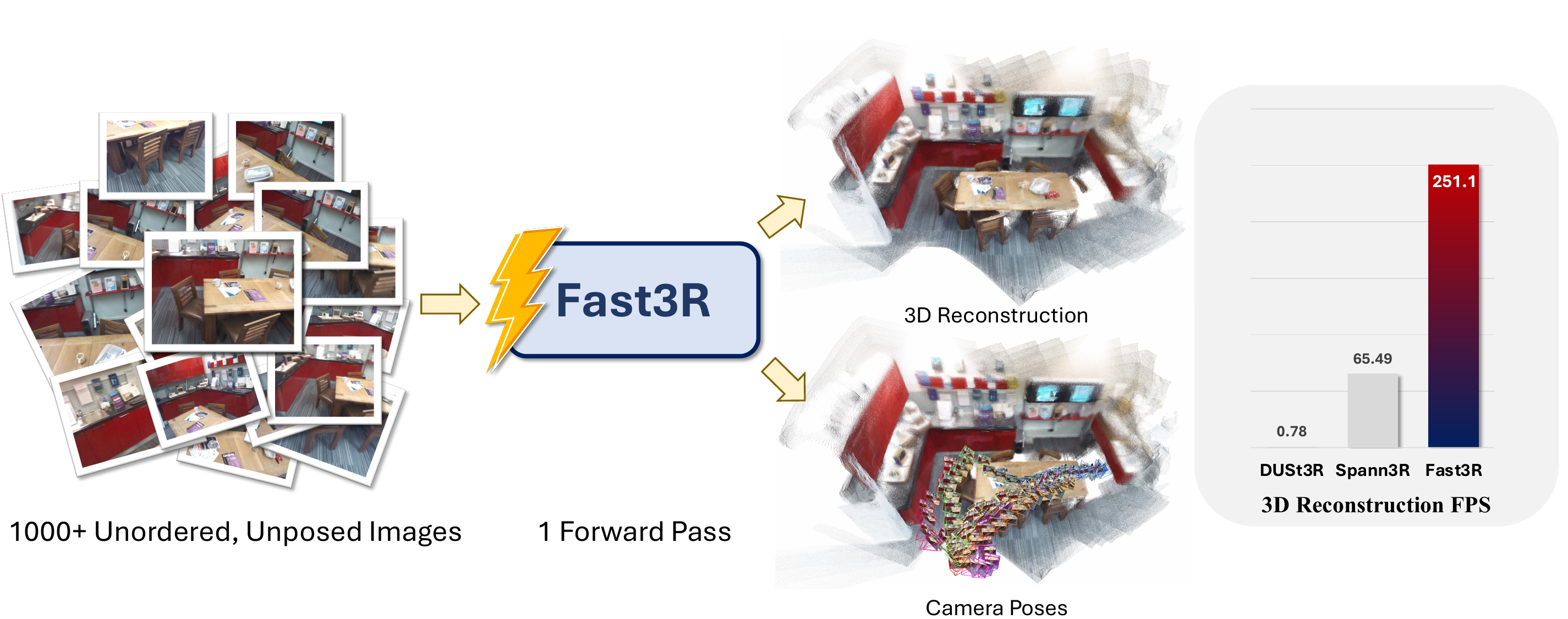}
    \captionof{figure}{\textbf{Fast3R} is a method towards 3D reconstructing 1000+ unordered, unposed images in a single forward pass. 
    }
    \vspace{0.6cm}
}]

\begin{abstract}
Multi-view 3D reconstruction remains a core challenge in computer vision, particularly in applications requiring accurate and scalable representations across diverse perspectives. Current leading methods such as DUSt3R employ a fundamentally pairwise approach, processing images in pairs and necessitating costly global alignment procedures to reconstruct from multiple views. 
In this work, we propose Fast 3D Reconstruction (\textbf{Fast3R}), a novel multi-view generalization to DUSt3R that achieves efficient and scalable 3D reconstruction by processing many views in parallel. Fast3R's Transformer-based architecture forwards $N$ images in a single forward pass, bypassing the need for iterative alignment.
Through extensive experiments on camera pose estimation and 3D reconstruction, Fast3R demonstrates state-of-the-art performance, with significant improvements in inference speed and reduced error accumulation. These results establish Fast3R as a robust alternative for multi-view applications, offering enhanced scalability without compromising reconstruction accuracy.
\end{abstract}

\section{Introduction}
\label{sec:intro}
3D reconstruction from multiple views has long been a foundational task across applications in autonomous navigation, augmented reality, and robotics~\cite{thrun2002probabilistic, mur2015orb}. Establishing correspondences across images, known as \textit{multi-view matching}, is central to these applications and enables an accurate scene representation. Traditional reconstruction pipelines, such as those based on Structure-from-Motion (SfM)~\cite{schoenberger2016sfmColmap} and Multi-View Stereo (MVS)~\cite{galliani2015massively}, fundamentally rely on image pairs to reconstruct 3D geometry. While effective in some settings, these methods require extensive engineering to manage the sequential stages of feature extraction, correspondence matching, triangulation, and global alignment, limiting scalability and speed.

This traditional ``pipeline'' paradigm has recently been challenged by DUSt3R~\cite{dust3r_cvpr24}, which directly predicts 3D structure from RGB images.
It achieves this with a design that ``cast[s] the pairwise reconstruction problem as a regression of pointmaps, relaxing the hard constraints of usual projective camera models"~\cite{dust3r_cvpr24}, yielding impressive robustness across challenging viewpoints.
This represents a radical shift in 3D reconstruction, as an end-to-end learnable solution is less prone to pipeline error accumulation, while also being dramatically simpler.

On the other hand, a fundamental limitation of DUSt3R is its restriction to two image inputs.
While image pairs are an important use case, often one is interested in reconstructing from more than two views, as when scanning of objects~\cite{reizenstein2021common} or scenes~\cite{dai2017scannet, yeshwanth2023scannet++, baruch2021arkitscenes, EPICFields2023, grauman2024egoexo4d}, \eg for asset generation or mapping.
To process more than two images, DUSt3R computes $\mathcal{O}(N^2)$ pairs of pointmaps and performs a global alignment optimization procedure.
This process can be computationally expensive, scaling poorly as the collection of images grows.
For instance, it will lead to OOM with only 48 views on an A100 GPU. 

Moreover, such a process is still fundamentally pairwise, which limits the model's context, both affecting learning during training and ultimate accuracy during inference.
In this sense, DUSt3R suffers from the same pair-wise bottleneck as traditional SfM and MVS methods.

We propose \textbf{Fast3R}, a novel multi-view reconstruction framework designed to overcome these limitations. Building on DUSt3R's foundations, Fast3R leverages a Transformer-based architecture~\cite{vaswani2017attention} that processes multiple images in parallel, allowing $N$ images to be reconstructed in a single forward pass. By eliminating the need for sequential or pairwise processing, each frame can simultaneously attend to all other frames in the input set during reconstruction, significantly reducing error accumulation. Perhaps surprisingly, \model{} also takes significantly less time.  

\noindent Our contributions are threefold.
\begin{enumerate}
    \item We introduce \model{}, a Transformer-based model for multi-view pointmap estimation that obviates the need for global postprocessing; resulting in significant improvements in speed, computation overhead and scalability.
    \item We show empirically that the model performance improves by scaling along the view axis. For camera pose localization and reconstruction tasks, the model improves when trained on progressively larger sets of views. Per-view accuracy further improves when more views are used during inference, and the model can generalize to significantly more views than seen during training. 
    \item We demonstrate state-of-the-art performance in camera pose estimation with significant inference time improvements. On CO3Dv2~\cite{reizenstein2021common}, \model{} gets \textbf{99.7\%} accuracy within 15-degrees for pose estimation, over a \textbf{14x} error reduction compared to DUSt3R \emph{with} global alignment.
\end{enumerate}

\model{} offers a scalable and accurate alternative for real-world applications, setting a new standard for efficient multi-view 3D reconstruction.

\begin{figure*}
    \centering
    \includegraphics[width=\linewidth]{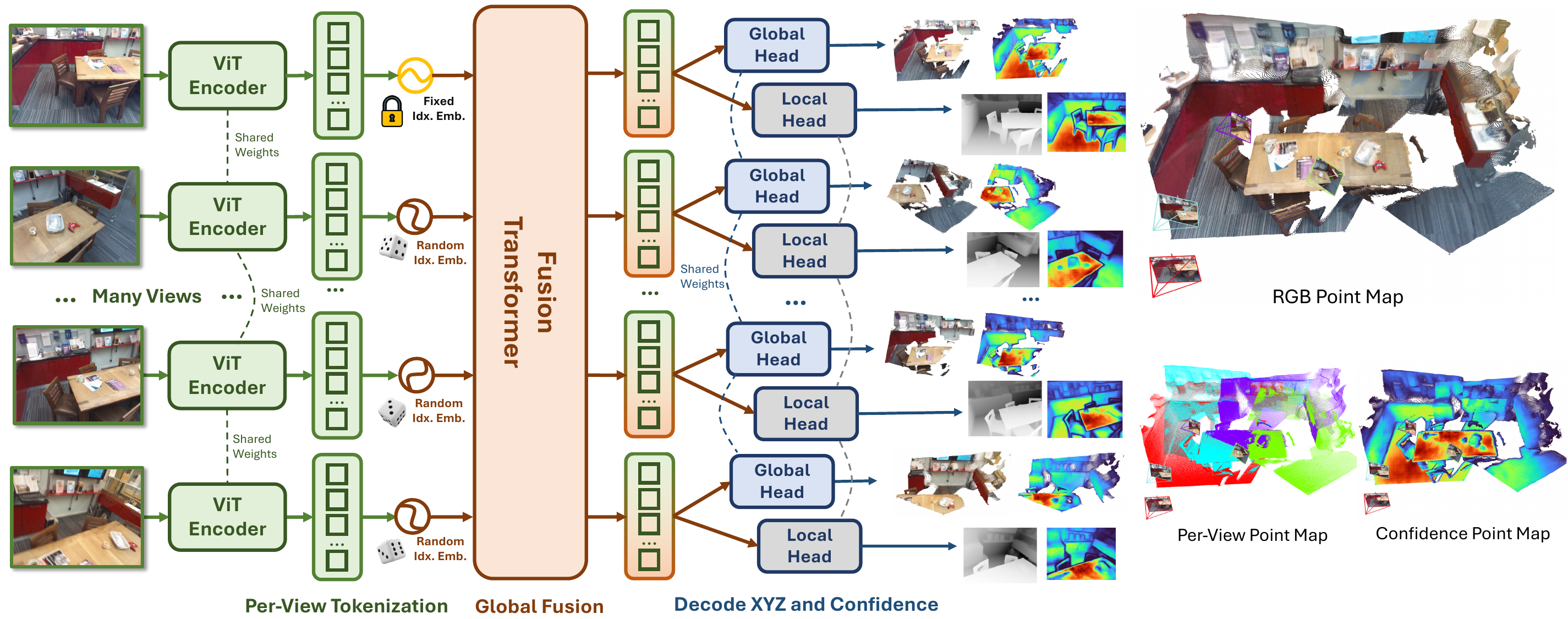}
    \caption{\textbf{Model architecture of Fast3R}.
    Built upon a novel Transformer-based architecture which supports bidirectional information flow, Fast3R is able to process dense input views simultaneously. }
    \label{fig:architecture}
\end{figure*}

\section{Related Work}
\label{sec:related_work}

\noindent\textbf{Multi-view 3D reconstruction:}
Almost all modern 3D reconstruction approaches are based on the traditional multi-view geometry (MVG) pipeline~\cite{HartleyZisserman2003MVG}. MVG-based methods first identify corresponding pixels between image pairs, and then use camera models and projective multiview geometry to lift these correspondences to 3D points. The process happens in sequential stages: feature extraction, finding pairwise image correspondences, triangulation to 3D and pairwise relative camera pose, and global bundle alignment. 
However, any pipeline approach is prone to accumulating errors, which are especially common in hand-crafted components. Moreover, the sequential nature prevents parallelization, which limits speed and scalability. 

MVG approaches have existed since the early days of computer vision, and are still in use for a reason: they can be highly accurate when they do not catastrophically fail. 
The latest multi-view geometry pipelines like COLMAP~\cite{schoenberger2016sfmColmap} or OrbSLAM2~\cite{murORB2} incorporate nearly 60 years of compounding engineering improvements, but these approaches still catastrophically fail $>$40\% of the time on static scenes like ETH-3D~\cite{teed2021droid}), which can actually be considered an easy case due to dense image coverage of the scene.

Much recent work has successfully addressed the robustness and speed by replacing increasingly large components of MVG pipelines with end-to-end learned versions that are faster and reduce the rate of catastrophic failures~\cite{Wang2023VGGSfMVG, Smith2024FlowMapHC, Zhao2022ParticleSfMED}. For example, ~\cite{sarlin20superglue, Gleize_2023_ICCV_silk, mast3r_arxiv24, Dusmanu2019D2NetAT, Yi2016LIFTLI, Teed2021DROIDSLAMDV} improve feature extraction and correspondences,~\cite{Tang2024ADen, lin2023relpose++, Wang2023PoseDiffusionSP, Zhang2024CamerasAR} learn to estimate camera pose, and~\cite{teed2021droid} introduce a bundle adjustment layer. ~\cite{dust3r_cvpr24} contains an excellent and comprehensive survey of such efforts. Overall, the trend is towards replacing increasingly large components with end-to-end solutions.

\noindent\textbf{Pointmap representation:} DUSt3R~\cite{dust3r_cvpr24} takes this evolution the furthest by proposing \emph{pointmap} regression to replace everything in the MVG pipeline up to global pairwise alignment. Rather than first attempting to solve for camera parameters in order to triangulate corresponding pixels, DUSt3R trains a model to directly predict 3D pointmaps for pairs of images in a shared coordinate frame.
Other MVG component tasks such as relative camera pose estimation and depth estimation can be recovered from the resulting pointmap representation.
However, DUSt3R's pairwise assumption is a limitation, as it requires inference on $\mathcal{O}(N^2)$ image pairs and then a global alignment optimization, which is per-scene and does not improve with more data. Moreover, this process quickly becomes slow or crashes due to exceeded system memory, even for relatively modest numbers of images.

DUSt3R has inspired several follow-ups. 
MASt3R~\cite{mast3r_arxiv24} adds a local feature head to each decoder's output, while MonST3R~\cite{zhang2024monst3r} does a data-driven exploration of dynamic scenes, but both are still fundamentally pairwise methods. 
MASt3R in particular does not make any changes to the global alignment methodology. 
Concurrently with our work, Spann3R~\cite{wang2024spann3r} treats images as an ordered sequence (\eg from a video) and incrementally reconstructs a scene using a pairwise sliding window network, along with a learned spatial memory system. This extends DUSt3R to handle more images, but Spann3R's incremental pairwise processing cannot fix reconstructions from earlier frames, which can cause errors to accumulate. Crucially, \model{}'s transformer architecture uses all-to-all attention, allowing the model to reason simultaneously and jointly over all frames without any assumption of image order. \model{} removes sequential dependencies, enabling parallelized inference across many devices in a single forward pass. 
\begin{figure*}[t]
    \centering
    \includegraphics[width=\textwidth]{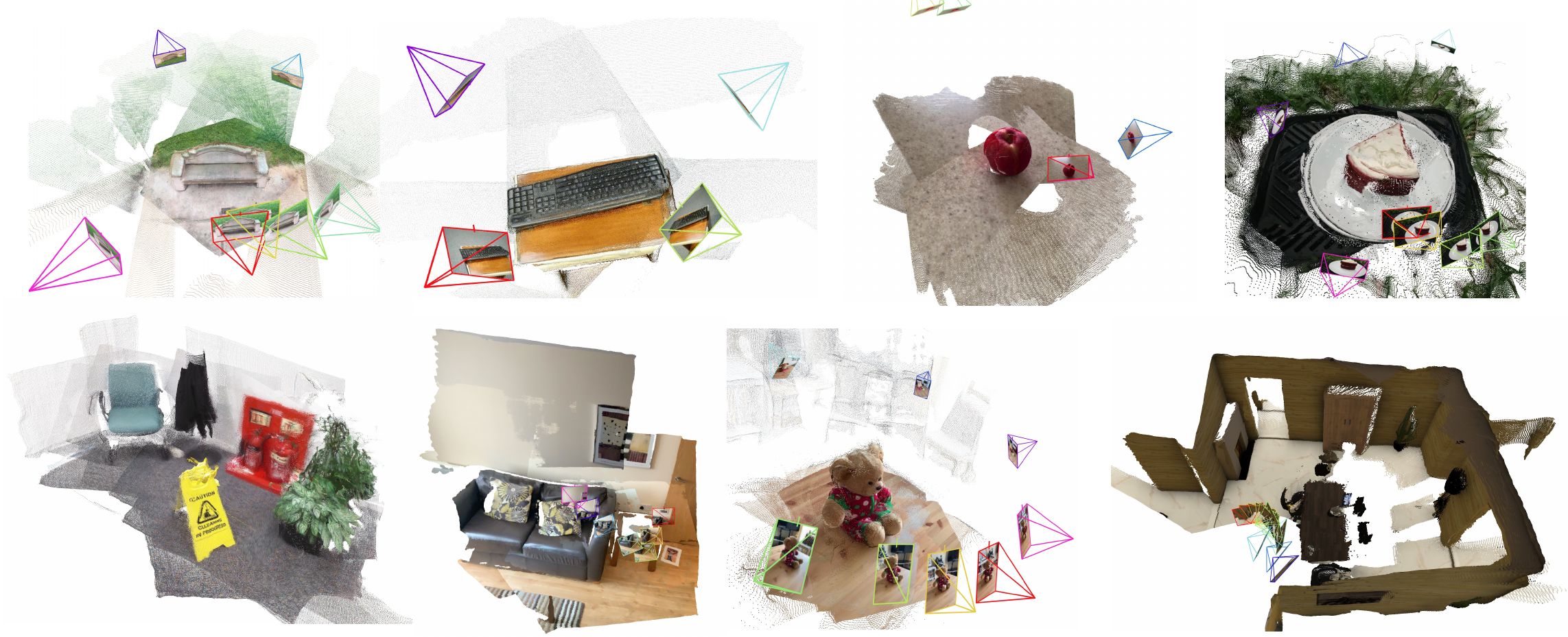}
    \vspace{-2mm}
    \caption{\textbf{Qualitative examples of \model{}'s output.} The text on the yellow sign says ``Caution, cleaning in progress" and is legible if zoomed in.}
    \label{fig:qualitative_examples}
    \vspace{-20pt}
\end{figure*}

\section{Model}

\model{} is a transformer-based model that predicts a 3D pointmap from a \emph{set} of unordered and unposed images. The model architecture is designed to be scalable to over 1000 images during inference, though during training we use image masking to train it with far fewer.
In this section, we detail our implementation of \model{}, and discuss the design choices that enable its scalability.

\subsection{Problem definition}
\label{sec:problem_definition}
Taking a set of ($N$) unordered and unposed RGB images $\mathbf{I} \in \mathbb{R}^{N \times H \times W \times 3}$ as inputs\footnote{We assume all images are resized to the same $H\times W$ for simplicity.}, \model{} reconstructs the 3D structures of the scene by predicting the corresponding \emph{pointmap} $\mathbf{X}$, where $\mathbf{X} \in \mathbb{R}^{N \times H \times W \times 3}$.
\emph{A pointmap is a set of 3D locations indexed by pixels} in an image $\mathbf{I}$, enabling the derivation of camera poses, depths, and 3D structures.

\model{} predicts two pointmaps: local pointmap $\mathbf{X}_L$ and global pointmap $\mathbf{X}_G$, and corresponding confidence maps $\Sigma_L$ and $\Sigma_G$ (of shape $\Sigma \in \mathbb{R}^{N \times H \times W}$). Overall, \model{} maps $N$ RGB images to $N$ local and global pointmaps:
\[
\mathrm{\model{}}: \mathbf{I} \rightarrow (\mathbf{X}_\text{L}, \Sigma_\text{L}, \mathbf{X}_\text{G}, \Sigma_\text{G})
\]
\noindent The global pointmap $\mathbf{X}_G$ is in the coordinate frame of the first camera and the $\mathbf{X}_L$ is in the coordinate frame of the viewing camera, as shown in Figure~\ref{fig:architecture}

\subsection{Training Objective}
This section describes the loss, using the notation in Section~\ref{sec:problem_definition} above. \model{}'s predictions of $(\hat{\mathbf{X}}_\text{L}, \hat{\Sigma_\text{L}}, \hat{\mathbf{X}}_\text{G}, \hat{\Sigma_\text{G}})$ are trained using generalized versions of the pointmap loss in DUST3R~\cite{dust3r_cvpr24}. 

Our total loss is the combination of pointmap losses for the local and global pointmaps:
\begin{equation}
\mathcal{L_{\text{total}}} = \mathcal{L_{\mathbf{X}_\text{G}}} + \mathcal{L_{\mathbf{X}_\text{L}}}
\end{equation}
which are confidence-weighted versions of the normalized 3D pointwise regression loss.

\noindent\textbf{Normalized 3D pointwise regression loss:}
The normalized regression loss for $\mathbf{X}$ is a multi-view version of that in DUSt3R~\cite{depthanything2_2024} or monocular depth estimation~\cite{depthanything2_2024, eftekhar2021omnidata, Ranftl2020midas}. It is the $L_2$ loss between the normalized predicted pointmaps and normalized target pointmaps, rescaled by the mean Euclidean distance to the origin:
\begin{equation}
\ell_{\text{regr}}(\hat{\mathbf{X}}, \mathbf{X}) = \left\lVert \frac{1}{\hat{z}} \hat{\mathbf{X}} - \frac{1}{z} \mathbf{X} \right\rVert _2~,~~~~~~~ z = \frac{1}{|\mathbf{X}|} \sum_{x \in \mathbf{X}} \lVert x \rVert_2
\end{equation}
Note that the predictions and targets are independently normalized by the mean euclidean distance to the origin.

\noindent\textbf{Pointmap loss:}
As in~\cite{dust3r_cvpr24}, we use a confidence-adjusted version of the loss above, using the confidence score $\hat\Sigma$ predicted by the model.
The total loss for a pointmap is
\begin{equation}
\mathcal{L_{\mathbf{X}}}(\hat{\Sigma}, \hat{\mathbf{X}}, \mathbf{X}) = \frac{1}{|\mathbf{X}|} \sum\hat \Sigma_{+} \cdot \ell_{\text{regr}}(\hat{\mathbf{X}}, \mathbf{X}) + \alpha \log(\hat\Sigma_{+})
\end{equation}
Since the $\log$ term requires the confidence scores to be positive, we enforce $\hat\Sigma_{+} = 1 + \exp(\hat\Sigma)$.
Our intuition is that the confidence weighting helps the model deal with label noise. Like DUSt3R, we train on real-world scans typically containing systematic errors in the underlying pointmap labels. For example, glass or thin structures are often not reconstructed properly in the ground-truth laser scans~\cite{yeshwanth2023scannet++,baruch2021arkitscenes}, and errors in camera registration will cause misalignments between the images and pointmap labels~\cite{depthanything2_2024}.

\subsection{Model architecture}

The \model{} meta-architecture is inspired by DUSt3R, and has three components: \emph{image encoding}, \emph{fusion transformer}, and \emph{pointmap decoding}, as shown in Figure~\ref{fig:architecture}. We emphasize that \model{} makes no assumptions on the ordering of images in $\mathbf{I}$, and all output pointmaps and confidence maps  $(\mathbf{X}_\text{L}, \Sigma_\text{L}, \mathbf{X}_\text{G}, \Sigma_\text{G})$ are predicted simultaneously, not sequentially.

\noindent \textbf{Image Encoder:}
\model{} encodes each image $I_i \in \mathbf{I}$ to a set of patch features $H_i$,  using a feature extractor $\mathcal F$.
This is done independently per image, yielding a sequence of image patch features $H_i = \{h_{i,j}\}_{j=1}^{HW/P^2}$ for each image:
\begin{equation}
    H_i = \mathcal{F}(I_i), i \in {1, ..., N}
\end{equation}
We follow DUSt3R's design and use CroCo ViT~\cite{weinzaepfel2022croco} as our encoder, though we found DINOv2~\cite{oquab2023dinov2} works similarly. 

Before passing image patch features $\mathbf{H}$ to the fusion transformer, we add position embeddings with one-dimensional \emph{image index positional embeddings}. 
 
Index embeddings help the fusion transformer determine which patches come from the same image and are the mechanism for identifying $I_1$, which importantly defines the global coordinate frame. 
This is critical for allowing the model to implicitly reason about camera pose jointly for all images from an otherwise permutationally invariant set of tokens.

\noindent\textbf{Fusion Transformer:}
Most of the computation in \model{} happens in the fusion transformer. We use a 24-layer transformer similar to ViT-L~\cite{dosovitskiy2021an}. This fusion transformer takes the concatenated encoded image patches from \textit{all} views and performs all-to-all self-attention. This operation provides \model{} with full context from \emph{all views}, beyond the information provided in pairs alone.


\noindent\textbf{Pointmap Decoding Heads:}
Finally, \model{} uses two separate DPT~\cite{ranftl2021dpt} decoding heads to map these tokens to local and global pointmaps $(\mathbf{X}_\text{L}, \mathbf{X}_\text{G})$, and confidence maps $(\Sigma_L, \Sigma_G$).

\noindent\textbf{Image index positional embedding generalization:} \label{sec:posembed-img-idx}
We would like \model{} to be able to handle many views at inference, more than were used to train a model. A na\"{i}ve way to embed views during testing would be to embed them in the same way as training: i.e. use the same Spherical Harmonic frequencies ~\cite{tancik2020fourier} to embed raw indices $SH(\{1,...,N\})$ during training, and $SH(\{1,...,N_{\mathrm{test}}\})$ during inference. In LLMs this causes poor performance, and in preliminary experiments, we also found that the resulting model did not work well when the number of input images exceeded that used during training (Sec. \ref{sec:posembed-img-idx-exps}). We therefore adopt Position Interpolation~\cite{chen2024extending}, a solution from LLMs, where during training we randomly draw $N$ indexes from a larger pool $N'$ of possible samples.~\cite{chen2024extending} draws samples using a regular grid since the LLM inputs form a regular ordered sequence. Our images are unordered, so we draw $N \subset \{1,...,N'\}$ uniformly at random. To the transformer, the strategy looks indistinguishable from masking out images, and $N' \gg N$ controls the masking ratio.\footnote{Patches $H_1$ from the first image $I_1$ are always embedded with $p_1$, since that image provides the coordinate frame for the global head.} This strategy enables \model{} to handle $N=1000$ images during inference, even if only trained with $N=20$ images.

\subsection{Memory-Efficient Implementation}

With a standard transformer architecture and a single-pass inference procedure, Fast3R is able to leverage many of the recent advances designed to improve scalability at train and inference time~\cite{jiang2024mixtralexperts, openai2024gpt4technicalreport, touvron2023llama2openfoundation, dubey2024llama3herdmodels}. 

For example, model size and throughput can be increased by sharding the model and/or data minibatch across multiple machines, such as through model parallelism~\cite{mesh-tf, gpipe}, data parallelism~\cite{pytorch-distributed}, and tensor parallelism~\cite{megatron-lm, megatron-lm2}. 
During training, optimizer weights, states, and gradients can also be sharded~\cite{deepspeed-zero}. Systems-level advances have also been proposed, such as FlashAttention~\cite{dao2022flashattention, dao2023flashattention2}, which uses highly optimized GPU kernels leveraging the hardware topology to compute attention in a time and memory-efficient way. These are implemented in libraries such as FAIRScale~\cite{FairScale2021}, DeepSpeed~\cite{deepspeed-zero} and HuggingFace~\cite{hf-transformers}, and require significant engineering effort.  

The \model{} meta-architecture is explicitly designed to take advantage of these efforts. We leverage two different forms of parallelism at training and inference time, as well as FlashAttention, described in more detail in Sec.~\ref{sec:exp}. More broadly, we believe that our approach will continue to benefit in the longer term as transformer-based scaling infrastructure continues to mature.

\begin{table*}[t!]
    \begin{center}
    \small
    \label{tab:relpose_mvs}
    \resizebox{0.82\linewidth}{!}{
    \begin{tabular}{lccccc c cc c}
    \toprule
    \multirow{2}{*}{Methods}  & \multicolumn{5}{c}{CO3Dv2~\cite{reizenstein2021common}} & & RealEstate10K~\cite{realestate10k} & & \multirow{2}{*}{FPS} \\ 
    \cline{2-6} \cline{8-8} 
                               & RRA@15$\uparrow$ & RRA@5$\uparrow$ & RTA@15$\uparrow$ & RTA@5$\uparrow$ & mAA(30)$\uparrow$ & & mAA(30)$\uparrow$ &  & \\ 
    \midrule
    Colmap+SG~\cite{DeTone2017SuperPointSI, Sarlin2019SuperGlueLF} & 36.1    & 24.4   & 27.3   & 17.2  & 25.3 & & 45.2 & & 0.056   \\
    PixSfM~\cite{Lindenberger2021PixelPerfectSW}                  & 33.7    & 26.1   & 32.9   & 17.6  & 30.1 & & 49.4 & & -    \\
    RelPose~\cite{Zhang2022RelPosePP}              & 57.1    & -      & -      & -     & -    & & -    & & 0.02    \\
    PosReg~\cite{Wang2023PoseDiffusionSP}           & 53.2    & -      & 49.1   & -     & 45.0 & & -    & & 0.015    \\
    PoseDiff~\cite{Wang2023PoseDiffusionSP}   & 80.5    & 59.5   & 79.8   & 61.7  & 66.5 & & 48.0 & & 0.015    \\
    RelPose++~\cite{lin2023relpose++}           & (85.5)  & -      & -      & -     & -    & & -    & & 0.02    \\
    DUSt3R~\cite{Wang2023DUSt3RG3}             & 96.2    & -      & 86.8   & -     & 76.7 & & 67.7 & & 0.78    \\
    MASt3R~\cite{leroy2024grounding}            & 94.6    & 93.2   & \bf{91.9}   & \bf 86.2  & 81.8 & & \textbf{76.4} & & 0.23   \\ 
    \bf{Fast3R-no-outdoor (Ours)}  &  \bf 99.7 & \bf 97.4      & 87.1   & 76.1 & \bf{82.5} & & - & & \bf 251.1    \\ 
    \bf{Fast3R (Ours)}  &  96.2 & 90.2      & 81.6 & 68.2 & 75.0 & & 72.7 & & \bf 251.1    \\ 
    \bottomrule
    \end{tabular}
}
\caption{\textbf{Multi-view pose regression on the CO3D~\cite{reizenstein2021common} and RealEstate10K~\cite{realestate10k} datasets}. 
        Parentheses denote methods that do not report results on the 10 views set; we report their best for comparison (8 views). \model{} does not assume known camera intrinsics. 
}
    \label{fig:cam_pose}
\normalsize
\end{center}
\vspace{-7mm}
\end{table*}

\begin{table}[t]
  \centering
  \footnotesize
  \setlength{\tabcolsep}{0.3em}
    \begin{tabularx}{\columnwidth}{>{\centering\arraybackslash}X >{\centering\arraybackslash}X >{\centering\arraybackslash}X >{\centering\arraybackslash}X >{\centering\arraybackslash}X}
      \toprule
         & \multicolumn{2}{c}{\textbf{Fast3R}} & \multicolumn{2}{c}{\textbf{DUSt3R}} \\
         \cmidrule(lr){2-3} \cmidrule(lr){4-5}
         \# Views & Time (s) & Peak GPU Mem (GiB) & Time (s) & Peak GPU Mem (GiB) \\
      \midrule
         2  & 0.065 & 3.84 & 0.092 & 3.52 \\
         8  & 0.122 & 6.33 & 8.386 & 24.59 \\
         32 & 0.509 & 13.25 & 129.0 & 67.61 \\
         48 & 0.84 & 20.8 & OOM & OOM \\
         320 & 15.938 & 41.90 & OOM & OOM \\
         800 & 89.569 & 55.97 & OOM & OOM \\
         1000 & 137.62 & 63.01 & OOM & OOM \\
         1500 & 308.85 & 78.59 & OOM & OOM \\
     \bottomrule
    \end{tabularx}
    \vspace{-5pt}

    \caption{\textbf{System performance metrics for different view counts on Fast3R and DUSt3R on a single A100.} Time is measured in seconds (s), and memory is measured in gibibytes (GiB). Each view is 512x384 in resolution. For DUSt3R, 
    at 48 views the $N^2$ pairwise reconstructions eventually consume all VRAM at its global alignment stage. Note that \model{}'s reported fastest FPS of 251.1 uses 108 views in 224x224 resolution.}
    \vspace{-5pt}
    \label{tab:system_usage_by_n_views}
\end{table}

\section{Experiments}
\label{sec:exp}

\noindent\textbf{Training Data}
\label{sec:datasets}
We train on a mix of real-world object-centric and scene scan data: CO3D~\cite{reizenstein2021common}, ScanNet++~\cite{yeshwanth2023scannet++}, ARKitScenes~\cite{baruch2021arkitscenes}, Habitat~\cite{habitat}, BlendedMVS~\cite{yao2020blendedmvs}, and MegaDepth~\cite{megadepth}. This is a subset of the datasets in DUSt3R, specifically 6 of the 9 datasets.

\noindent\textbf{Baselines}
DUSt3R~\cite{dust3r_cvpr24} is the closest approach to ours, and competitive on visual odometry and reconstruction benchmarks. That paper contains extensive comparisons against other methods, and we adopt it as our main baseline.
We additionally consider DUSt3R's follow-up work, MASt3R~\cite{mast3r_arxiv24}, as well as a concurrent work Spann3R~\cite{wang2024spann3r}, which also seeks to replace DUSt3R's expensive global alignment stage by sequentially processing frames with an external spatial memory.
For camera pose estimation and 3D reconstruction, we include comparisons to task-specific methods. 

\noindent\textbf{Architecture Details}
In our experiments, we use the following components for the meta-architecture:
\begin{enumerate}
    \item The \textit{Image Encoder} uses a ViT-Large~\cite{weinzaepfel2022croco} architecture, initialized with DUSt3R pretrained weights~\cite{dust3r_cvpr24}. The ViT-L uses 16x16 patch size, and has 24 layers, 16 heads, embedding dimension size 1024, and MLP ratio 4.0.
    \item The \textit{Fusion Transformer} is a ViT-Large model initialized from scratch. We use a pool size of $N'=1000$ for image index embeddings.
    \item The \textit{Pointmap Decoding Heads} include two heads: global and local, both of which are DPT heads following \cite{ranftl2021dpt}. We initialize the global head with DUSt3R pretrained weights, and initialize the local head from scratch.
\end{enumerate}

\noindent\textbf{Training Details}
Our models are trained on images with 512 resolution (512 on the longest side) using AdamW~\cite{adam-w} for 174K steps, with a learning rate of 0.0001 and cosine annealing schedule. Unlike DUSt3R, we do not use staged training.
We implement multi-view dataloaders that can load N images in each sample. We train with batch size 128, with each sample consisting of a tuple of $N=20$ views. This process takes 6.13 days on 128 Nvidia A100-80GB GPUs.
We additionally make use of several strategies to enable efficient training. First, we use the FlashAttention~\cite{dao2022flashattention, dao2023flashattention2} to improve time and memory efficiency.
Even so, a na\"ive implementation runs out of memory even with batch size 1 when $N>16$, so we use DeepSpeed ZeRO stage 2 training \cite{deepspeed-zero}, whereby optimizer states, moment estimates, and gradients are partitioned across different machines. This enables us to train with up to $N=28$ views per data sample at max, with a batch size of one per GPU.

\subsection{Inference Efficiency}
\label{sec:inference-efficiency}

At inference time, we aim to handle 1000+ views compared to 20 during training, which requires additional optimizations. We observe the memory bottleneck at inference is in the DPT heads producing the pointmaps: with 320 views on a single A100 GPU, over 60\% of VRAM is consumed by activations from the DPT heads,
largely due to each needing to upsample $1024$ tokens into a high-resolution $512\times512$ image. To address this, we implement a simple version of tensor parallelism, putting the model on GPU 0 and then copying the DPT heads to each of the $K-1$ other GPUs. When processing a batch of $N \approx 1000$ images, we pass the entire batch through the ViT encoder and global fusion decoder, and then split the outputs across $K$ machines for parallel DPT head inference.

Table \ref{tab:system_usage_by_n_views} shows the inference time and memory usage as we increase the number of views. \model{} is able to process up to 1500 views in a single pass, whereas DUSt3R runs out of memory past $32$. \model{} also has a significantly faster inference time, with gains that increase with more views.

\begin{table}[t]
  \centering
  \footnotesize
  \setlength{\tabcolsep}{0.3em}
    \begin{tabularx}{0.98\columnwidth}{r c >{\centering\arraybackslash}X >{\centering\arraybackslash}X >{\centering\arraybackslash}X >{\centering\arraybackslash}X}
      \toprule

           \multirow{2}{*}{Method} & \multirow{2}{*}{FPS} & \multicolumn{2}{c}{7 scenes~\cite{shotton2013scene}} & \multicolumn{2}{c}{NRGBD~\cite{azinovic2022neural}} \\
      \cmidrule(lr){3-4} \cmidrule(lr){5-6}
         & & {\tt{Acc}$\downarrow$} & {\tt{Comp}$\downarrow$} & {\tt{Acc}$\downarrow$} & {\tt{Comp}$\downarrow$} \\
      \midrule
      {F-Recon~\cite{xu2023frozenrecon}} &  \footnotesize{\textless}0.1 & 7.62 & 2.31 & 20.59 & 6.31 \\
      {DUSt3R$^\dagger$~\cite{dust3r_cvpr24}} & 0.78 & \bf 1.23 & 0.91 & \bf 2.51 & 1.03 \\
      {Spann3R~\cite{wang2024spann3r}} & 65.4 & 1.48 & \bf 0.85 & 3.15 & 1.10 \\
      \bf {Fast3R (Ours)} & \bf 251.1 & 1.58 & 0.93 & 3.40 & \bf 1.01 \\
      \bottomrule
    \end{tabularx}%
    \vspace{-5pt}

    \caption{\textbf{Quantitative reconstruction results on scene datasets:} The numbers indicate {\tt{median}} distance to GT points on 7-Scenes~\cite{shotton2013scene} and NRGBD~\cite{azinovic2022neural} datasets. These datasets contain video trajectories of 500-1500 frames, and we evaluate using the same frame skip as other baselines. For 7-Scenes this is \texttt{skip}=20, and NRGBD uses \texttt{skip}=40. DUSt3R$^\dagger$ indicates using DUSt3R's final weights on $224\times 224$ images, to fit within GPU memory. Distances are scaled $100\times$ to remove the leading {\tt{0.00}}.
    } 
    \vspace{-10pt}
    \label{tab:quant_recon_scene}
\end{table}

\begin{table}[t]
  \centering
  \footnotesize
  \setlength{\tabcolsep}{0.3em}
    \begin{tabularx}{0.98\columnwidth}{r c >{\centering\arraybackslash}X >{\centering\arraybackslash}X}
      \toprule

         \multirow{2}{*}{Method} & \multirow{2}{*}{Views} & \tt{Acc}$\downarrow$ & \tt{Comp}$\downarrow$ \\ 
      \cmidrule(lr){3-3} \cmidrule(lr){4-4}
         & & \tt{Med.} & \tt{Med.} \\
      \midrule
      
      {DUSt3R~\cite{dust3r_cvpr24}} & {\tt all/5} 
        & \bf 1.159 & 0.914 \\
      {DUSt3R$^\dagger$~\cite{dust3r_cvpr24}} & {\tt all/5} 
        & 1.297 & 1.002 \\
      {Spann3R \cite{wang2024spann3r}} & {\tt all/5} 
        & 2.268 & 1.295 \\
       \textbf{Fast3R (Ours)} & {\tt all/5} 
        & 1.706 & \bf 0.857 \\

      \bottomrule
    \end{tabularx}
    \vspace{-5pt}

    \caption{\textbf{Quantitative results on object-centric DTU~\cite{aanaes2016large} dataset.} Using a {\tt skip}=5 on trajectories of 49 frames.}
    \vspace{-10pt}
    \label{tab:quant_recon_object}
\end{table}

\subsection{Camera Pose Estimation}
We evaluate camera pose estimation on unseen trajectories from 41 object categories from CO3D~\cite{reizenstein2021common}. Following \cite{dust3r_cvpr24}, we sample 10 random views from each trajectory.

Inspired by DUSt3R~\cite{dust3r_cvpr24}, we estimate the \emph{focal length}, \emph{camera rotation}, and \emph{camera translation} from the predicted global pointmaps. We begin by initializing a set of random focal length guesses based on the image resolution, then use RANSAC-PnP to estimate the camera's rotation and translation based on the guessed focal lengths and the global pointmap. The count of outliers from RANSAC-PnP is used to score each guessed focal length (lower is better), and the best-scoring focal length is selected to compute the intrinsic and extrinsic camera matrices.

During RANSAC-PnP, we only use points with the top 15\% confidence scores predicted by Fast3R, ensuring efficient PnP processing and reducing outliers. If all images are known to originate from the same physical camera, we use the focal length estimated from the first view as the focal length for all cameras, as this initial estimate has been empirically found to be more reliable. Otherwise, we independently estimate the focal length for each input.
It is worth noting that the camera pose estimation process is parallelized using multi-threading, ensuring minimal wall-clock time. Even for hundreds of views, the process completes in just a few seconds on standard CPUs.

We report Relative Rotation Accuracy (RRA) and Relative Translation Accuracy (RTA) at a threshold of $15^{\circ}$, mean Average Accuracy (mAA) at threshold $30^{\circ}$, and model frames per second (FPS) in Table~\ref{fig:cam_pose}.
On CO3D, \model{} surpasses all other methods across the RRA and mAA metrics, achieving near-perfect RRA, while remaining competitive on RTA. Importantly, it is also orders of magnitude faster: $\mathbf{320\times}$ faster than DUSt3R and $\mathbf{1000\times}$ faster than MASt3R. Fast3R-no-outdoor is the same Fast3R model but trained without the BlendedMVS and MegaDepth datasets. We find these datasets help the model generalize to more diverse scenes (e.g., drone shots and large outdoor landmarks) but slightly hurt pose estimation performance on CO3D.

\begin{figure}[t]
    \centering
    \includegraphics[width=0.49\textwidth]{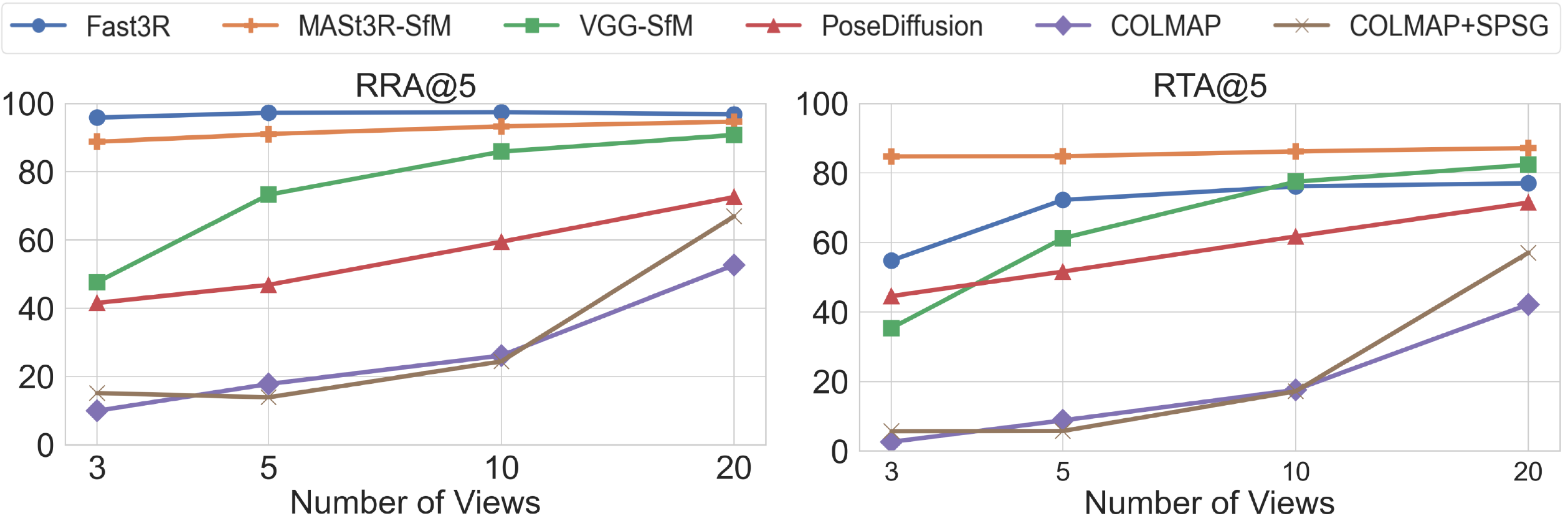}
    \caption{\textbf{Pose accuracy with more views:} \model{} improves with the context from more views. \model{} saturates the orientation portion of the benchmark, even using 3-5 views.}
    \label{fig:vo-scaling-inference-views}
    \vspace{-2mm}
\end{figure}

\begin{figure}
    \centering
    \includegraphics[width=0.9\linewidth]{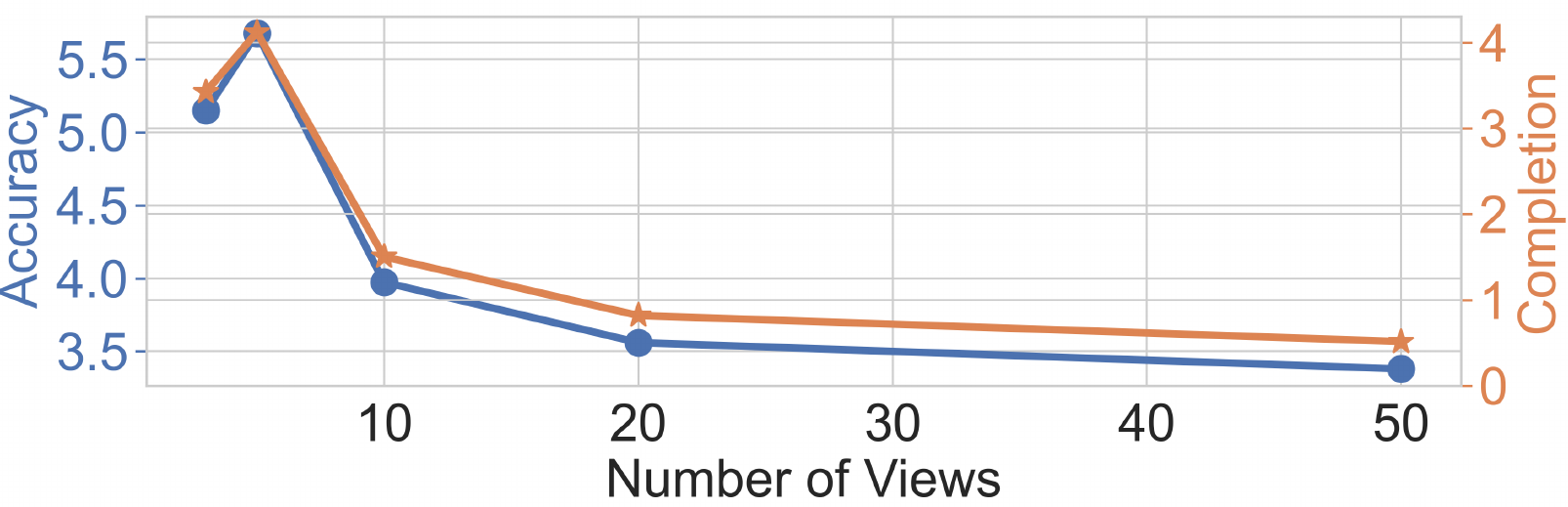}
    \vspace{-3mm}
    \caption{\textbf{DTU reconstruction quality vs. test number of views.} Accuracy and Completion (lower is better) get better as we inference with more views.}
    \label{fig:ablation_recon_vs_test_num_views}
    \vspace{-3mm}
\end{figure}

Figure~\ref{fig:vo-scaling-inference-views} and  Figure~\ref{fig:ablation_recon_vs_test_num_views} shows that \model{}'s predictions improve with more views, indicating that the model is able to use the additional context from multiple images.

\subsection{3D Reconstruction}
\label{sec:3d_reconstruction}
We evaluate \model{}'s 3D reconstruction on scene-level benchmarks: 7-Scenes~\cite{shotton2013scene} and Neural RGB-D~\cite{azinovic2022neural}, and the object-level benchmark DTU~\cite{aanaes2016large}. 

We found that local pointmap head learns more accurate pointmaps than the global pointmap head (ablation in Sec. \ref{sec:ablation_local_head}). Therefore we use the local pointmaps for detail and the global pointmaps for the high-level structure. Specifically, we independently align each image's local pointmap to the global pointmap using ICP and use aligned local pointmaps for evaluation. 

\model{} is competitive with other pointmap reconstruction methods like DUSt3r and MASt3R, while being significantly faster, as shown in Table~\ref{tab:quant_recon_scene} and Table~\ref{tab:quant_recon_object}. We believe that \model{} will continue to improve with better reconstruction data, more compute, and better training recipes. We show supportive scaling experiments in Figure~\ref{sec:scaling-training-views}.

\begin{figure}[t]
    \centering
    \includegraphics[width=0.9\linewidth]{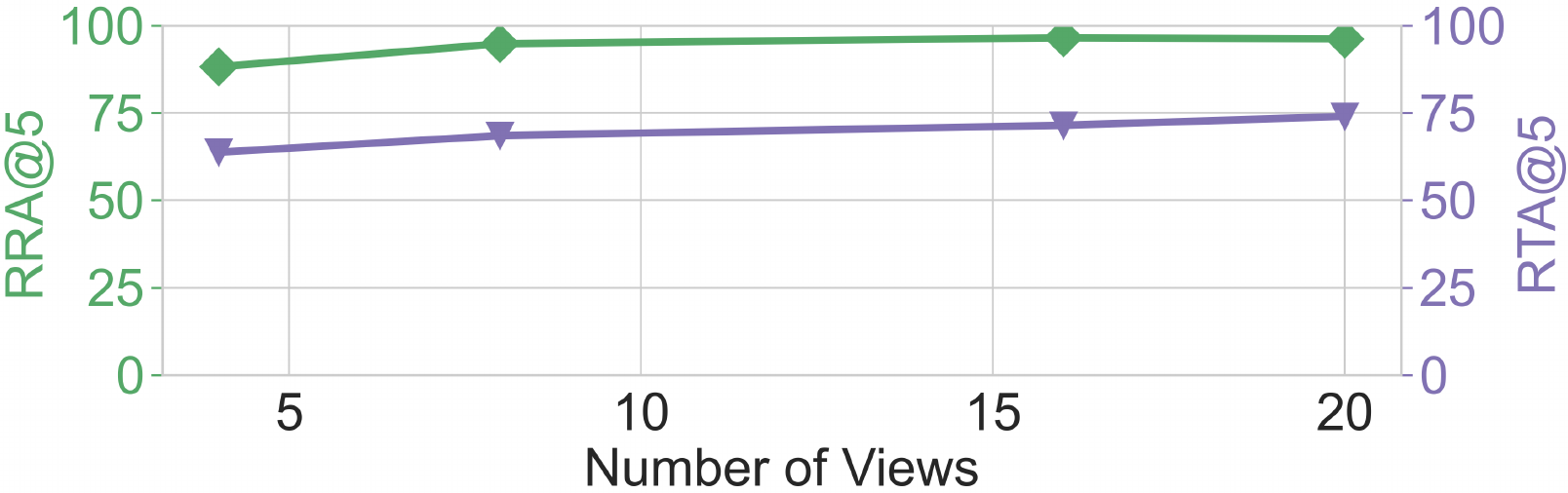}
    \caption{\textbf{Increasing \# views during training: camera pose estimation on CO3D.} Estimates of both orientation (RRA@5) and translation (RTA@5) improve with more views.}
    \label{fig:vo-scaling-train-views}
    \vspace{-20pt}
\end{figure}

\begin{figure}[t]
    \centering
    \includegraphics[width=0.4\textwidth]{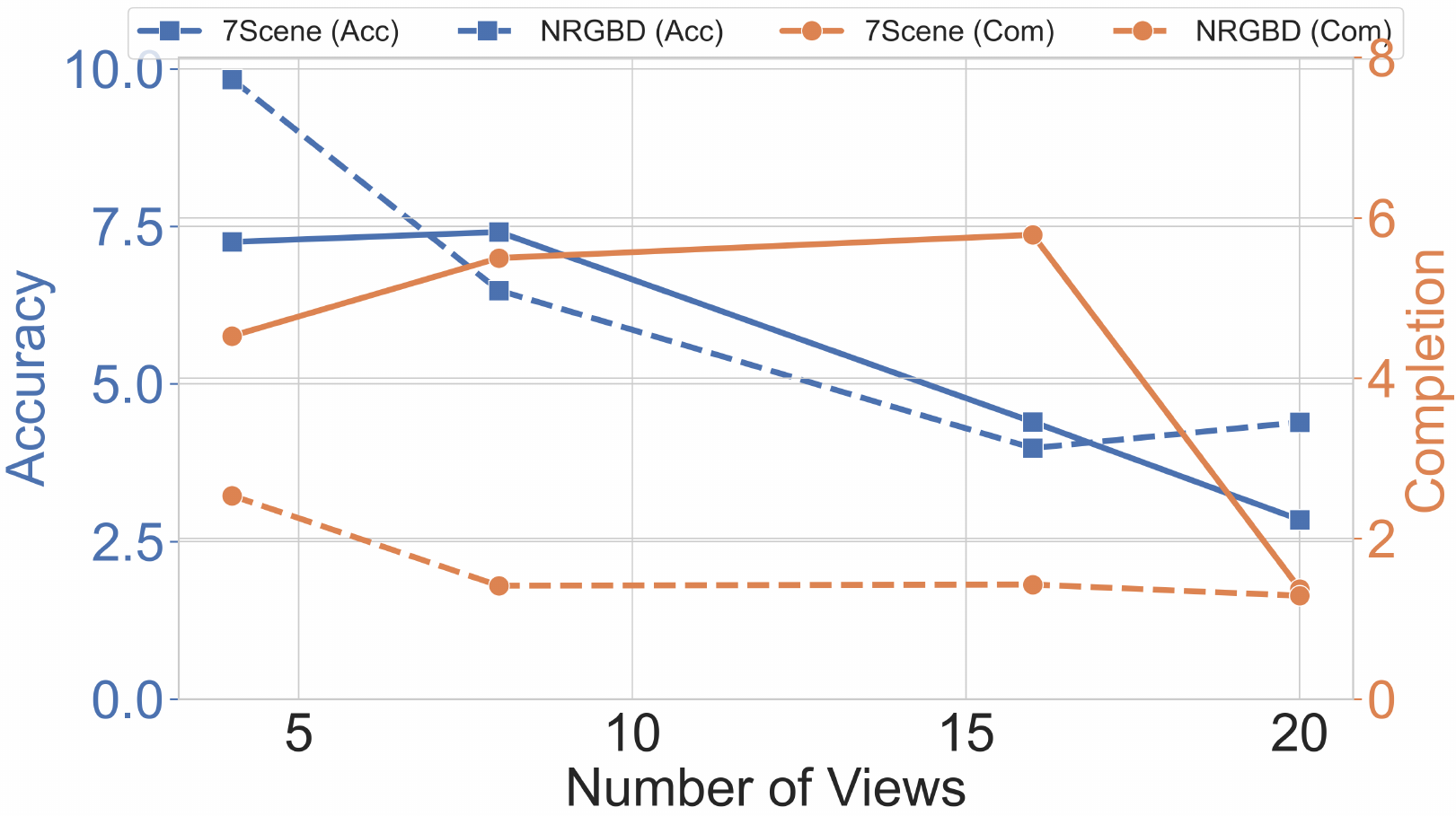}
    \caption{\textbf{Increasing \# views during training: reconstruction on 7scenes and NRGBD}. Accuracy and Completion (lower is better) get better as we train with more views. Normal Consistency (high is better) also gets better as we train with more views.}
    \label{fig:ablation_recon_vs_train_num_views}
\end{figure}

\section{Ablation Studies}
\label{sec:ablation}

\vspace{-5pt}
\subsection{Scaling the number of views}
\vspace{-5pt}
\label{sec:scaling-training-views}
\model{} is able to use all-to-all attention during training, which lets it learn from the global context. We hypothesize that the additional context provided by more views during training allows the model to learn higher-order correspondences between multiple frames, ultimately increasing model performance and increasing potential for scaling.

Figures~\ref{fig:vo-scaling-train-views} and \ref{fig:ablation_recon_vs_train_num_views} show that training on increasingly more views consistently improves RRA and RTA for visual odometry, and reconstruction accuracy|even when the number of views used during evaluation is held constant and the model is ultimately evaluated on \emph{fewer} views than were seen during training. 
We further evaluate the model's ability to reason about additional views by increasing the number of images that \model{} sees during inference. Figure~\ref{fig:vo-scaling-inference-views} and  Figure~\ref{fig:ablation_recon_vs_test_num_views} indicate that as the model uses more views, the average \emph{per-view} performance improves. This behavior holds for all evaluated metrics in both camera pose estimation and reconstruction. As shown in Figure~\ref{fig:ablation_recon_vs_test_num_views}, the model has a better per-view accuracy using 50 images than it does with 20, even though it was trained with 20. Many applications (\eg reconstruction, odometry) require inference on many views, which is a major motivation for \model{} removing the pairwise constraint. 

\vspace{-5pt}
\subsection{Model scaling and data scaling}
\vspace{-5pt}
The Transformer architecture's scalability is a key advantage of Fast3R. Although full experiments are in the Appendix \ref{sec:scaling_model} and \ref{sec:scaling_data}, our results show that increasing model size and data consistently boosts performance, promising better outcomes with more computational investment.

\begin{figure}[t]
    \centering
    \includegraphics[width=0.7\linewidth]{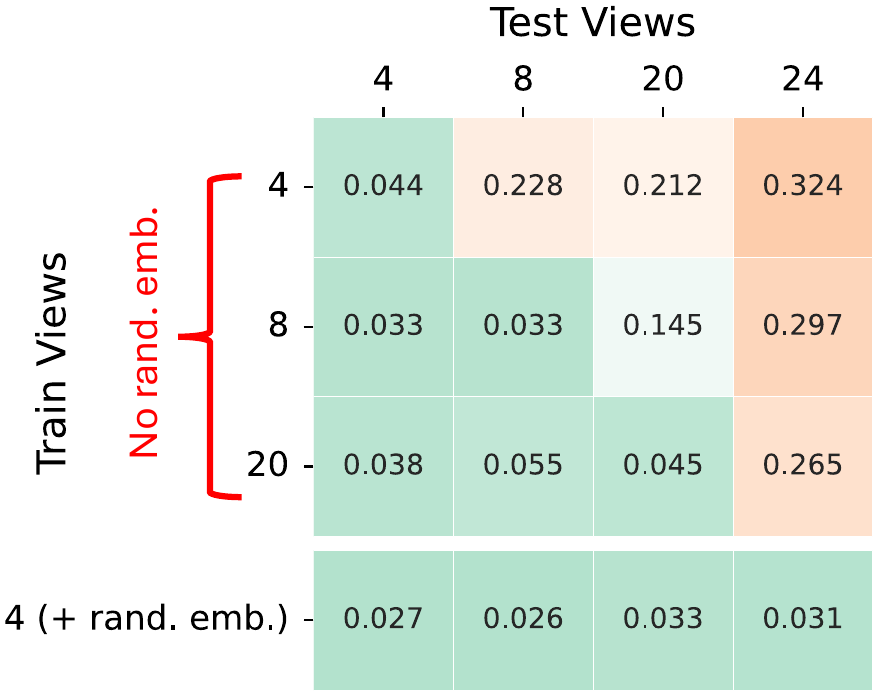}
    \vspace{-2mm}
    \caption{\textbf{Effect of sampling image index PE during training.} If we train the model without sampling index embeddings, regression loss spikes (\textcolor{orange}{orange}) when testing with more views than seen at training (top). Our embedding strategy performs comparably even with $6\times$ the number of views during training.}
    \label{fig:ablation_rand_emb}
    \vspace{-2mm}
\end{figure}

\vspace{-5pt}
\subsection{Training without position interpolation}
\vspace{-5pt}
\label{sec:posembed-img-idx-exps}
In Section~\ref{sec:posembed-img-idx}, we introduced a randomized version of~\cite{chen2024extending} to enable inference on more views than seen training. Without this technique, model accuracy quickly degrades for pointmap corresponding to image indexes outside the training range, as shown in Figure~\ref{fig:ablation_rand_emb} (top). A version of \model{} trained on $N=4$ views still produces high-quality pointmaps for views in slot 5 to 24 (Figure~\ref{fig:ablation_rand_emb} bottom).

\begin{table}[t]
    \centering
    \footnotesize
    \setlength{\tabcolsep}{0.3em}
    \begin{tabularx}{\linewidth}{r >{\centering\arraybackslash}X >{\centering\arraybackslash}X >{\centering\arraybackslash}X >{\centering\arraybackslash}X >{\centering\arraybackslash}X >{\centering\arraybackslash}X}
        \toprule
        \multirow{2}{*}{Pointmap Type} & \multicolumn{2}{c}{7-Scenes~\cite{shotton2013scene}} & \multicolumn{2}{c}{NRGBD~\cite{azinovic2022neural}} & \multicolumn{2}{c}{DTU} \\
        \cmidrule(lr){2-3} \cmidrule(lr){4-5} \cmidrule(lr){6-7}
        & {\tt{Acc}$\downarrow$} & {\tt{Comp}$\downarrow$} & {\tt{Acc}$\downarrow$} & {\tt{Comp}$\downarrow$} & {\tt{Acc}$\downarrow$} & {\tt{Comp}$\downarrow$} \\
        \midrule
        {local aligned to global} & 2.84 & 1.37 & 4.39 & 1.28 & 3.91 & 1.41 \\
        {global} & 4.81 & 1.64 & 4.85 & 1.32 & 3.88 & 1.41 \\
        {$\Delta$} & \textcolor{red}{+1.97} & \textcolor{red}{+0.27} & \textcolor{red}{+0.46} & \textcolor{red}{+0.04} & \textcolor{green}{-0.03} & 0.00 \\
        \bottomrule
    \end{tabularx}
    \vspace{-5pt}
    \caption{\textbf{Ablation on the effect of local head on 3D reconstruction.} \textcolor{red}{Red}/\textcolor{green}{green} indicate an increase/decrease in error compared to using the local pointmap aligned to the global pointmap.}
    \label{tab:ablation_local_head}
    \vspace{-10pt}
\end{table}

\vspace{-5pt}
\subsection{Inference without local head}
\vspace{-5pt}
\label{sec:ablation_local_head}
In Table~\ref{tab:ablation_local_head}, we perform an ablation experiment on the effect of using local vs. global pointmaps for 3D reconstruction (also see the visualization in Figure~\ref{fig:viz_local_alignment} in appendix). Specifically, we compare doing 3D reconstruction directly with the predicted global pointmaps to reconstruction with the predicted local pointmaps aligned to the global coordinate system using ICP (Sec. \ref{sec:3d_reconstruction}). Qualitative and quantitative results show that the local head produces more accurate pointmaps (fewer floaters, less smearing, less distortion) compared to the global head. We attribute this behavior to the local head being more invariant during training than the global head. That is, the local head is supervised such that the 3D XYZ locations of pixels in an image do not change, no matter which view is selected as the anchor view $I_1$; whereas for the global head, the XYZ locations for pixels in a view is dependent on which view is selected as the anchor view. Conceptually, the global head needs to learn both 2D-to-3D geometry and the rigid transformation of 3D points in the global coordinate system; whereas the local head only needs to learn 2D-to-3D geometry.

\section{Conclusion}
\vspace{-5pt}
We introduce \model{}, a transformer that predicts 3D locations for all pixels in a common frame of reference, directly in a single forward pass. By replacing the whole SfM pipeline with a generic architecture trained end-to-end, \model{} and similar approaches should benefit from the usual scaling rules for transformers: consistent improvement with better data and increased parameters. Since \model{} uses global attention, it avoids two potentially artificial scaling limits due to bottlenecks in existing systems. First, the bottleneck of image pair reconstruction restricts the information available to the model. Second, pairwise global optimization can only make up for this so much and does not improve with more data.

With our efficient implementation, \model{} can operate at $>250$ FPS, and process 1500 images in one forward pass, far exceeding other methods while achieving competitive results on 3D reconstruction and camera pose estimation benchmarks. We demonstrate that \model{} can be finetuned to reconstruct videos by changing the data and without modifying the pointmap regression objective and architecture.
In contrast with pipeline approaches bottlenecked by custom and slow operations, \model{} inherits the benefits of future engineering improvements to efficiently serve and train large transformer-based models. For example, packages like Deepspeed-Inference~\cite{deepspeed}, FlashAttention~\cite{dao2022flashattention, dao2023flashattention2} provide fused kernels, model parallelism, and data parallelism. These speed up inference and reduce memory requirements, allowing more images per device, and the number of images scales with the number of devices.

\noindent \textbf{Limitations:} A current limiting factor for scaling may be data accuracy and quantity. Synthetic data~\cite{infinigen, hypersim} could be a solution as, broadly speaking, models trained for geometry estimation seem to generalize well from simulation data. \model{} can successfully use simulated data to train for 4D reconstruction, showing generalization results on DAVIS. Similarly, DepthAnythingV2~\cite{depthanything2_2024} showed the potential of this approach to scale for monocular depth estimation.

The architecture of Fast3R allows for parallel processing of many views, and its positional embedding design enables ``train short, test long" in terms of context length of views. However, we observed that for scenes with very large reconstruction areas, when the number of views becomes extreme (e.g., more than 300), the point map of some views (particularly those with a low confidence score) begins to exhibit drifting behavior. One current way to address this issue is to drop frames with low confidence scores. In dense reconstruction, this approach typically does not hurt reconstruction quality too much. However, to fundamentally address this problem, we hypothesize that future work could explore the following avenues: (1) incorporating more data containing large scenes to improve generalization to such cases; (2) designing better positional embeddings inspired by state-of-the-art long-context language models \cite{zhupose}, which can handle very long context lengths and exploit the temporal structure of ordered image sequences (e.g., video).

{
    \small
    \bibliographystyle{ieeenat_fullname}
    \bibliography{main}

\begin{thebibliography}{76}
\providecommand{\natexlab}[1]{#1}
\providecommand{\url}[1]{\texttt{#1}}
\expandafter\ifx\csname urlstyle\endcsname\relax
  \providecommand{\doi}[1]{doi: #1}\else
  \providecommand{\doi}{doi: \begingroup \urlstyle{rm}\Url}\fi

\bibitem[Aan{\ae}s et~al.(2016)Aan{\ae}s, Jensen, Vogiatzis, Tola, and Dahl]{aanaes2016large}
Henrik Aan{\ae}s, Rasmus~Ramsb{\o}l Jensen, George Vogiatzis, Engin Tola, and Anders~Bjorholm Dahl.
\newblock Large-scale data for multiple-view stereopsis.
\newblock \emph{International Journal of Computer Vision}, 120:\penalty0 153--168, 2016.

\bibitem[Achiam et~al.(2024)Achiam, Adler, Agarwal, Ahmad, Akkaya, Aleman, Almeida, Altenschmidt, Altman, Anadkat, et~al.]{openai2024gpt4technicalreport}
Josh Achiam, Steven Adler, Sandhini Agarwal, Lama Ahmad, Ilge Akkaya, Florencia~Leoni Aleman, Diogo Almeida, Janko Altenschmidt, Sam Altman, Shyamal Anadkat, et~al.
\newblock Gpt-4 technical report, 2024.

\bibitem[Azinovi{\'c} et~al.(2022)Azinovi{\'c}, Martin-Brualla, Goldman, Nie{\ss}ner, and Thies]{azinovic2022neural}
Dejan Azinovi{\'c}, Ricardo Martin-Brualla, Dan~B Goldman, Matthias Nie{\ss}ner, and Justus Thies.
\newblock Neural rgb-d surface reconstruction.
\newblock In \emph{Proceedings of the IEEE/CVF Conference on Computer Vision and Pattern Recognition}, pages 6290--6301, 2022.

\bibitem[Baruch et~al.(2021)Baruch, Chen, Dehghan, Dimry, Feigin, Fu, Gebauer, Joffe, Kurz, Schwartz, and Shulman]{baruch2021arkitscenes}
Gilad Baruch, Zhuoyuan Chen, Afshin Dehghan, Tal Dimry, Yuri Feigin, Peter Fu, Thomas Gebauer, Brandon Joffe, Daniel Kurz, Arik Schwartz, and Elad Shulman.
\newblock Arkitscenes: A diverse real-world dataset for 3d indoor scene understanding using mobile rgb-d data.
\newblock \emph{arXiv preprint arXiv:2111.08897}, 2021.

\bibitem[Chen et~al.(2024)Chen, Wong, Chen, and Tian]{chen2024extending}
Shouyuan Chen, Sherman Wong, Liangjian Chen, and Yuandong Tian.
\newblock Extending context window of large language models via positional interpolation.
\newblock In \emph{Proceedings of the International Conference on Learning Representations (ICLR)}, 2024.

\bibitem[Dai et~al.(2017)Dai, Chang, Savva, Halber, Funkhouser, and Nie{\ss}ner]{dai2017scannet}
Angela Dai, Angel~X. Chang, Manolis Savva, Maciej Halber, Thomas Funkhouser, and Matthias Nie{\ss}ner.
\newblock Scannet: Richly-annotated 3d reconstructions of indoor scenes.
\newblock In \emph{Proc. Computer Vision and Pattern Recognition (CVPR), IEEE}, 2017.

\bibitem[Dao(2024)]{dao2023flashattention2}
Tri Dao.
\newblock Flash{A}ttention-2: Faster attention with better parallelism and work partitioning.
\newblock In \emph{International Conference on Learning Representations (ICLR)}, 2024.

\bibitem[Dao et~al.(2022)Dao, Fu, Ermon, Rudra, and R{\'e}]{dao2022flashattention}
Tri Dao, Daniel~Y. Fu, Stefano Ermon, Atri Rudra, and Christopher R{\'e}.
\newblock Flash{A}ttention: Fast and memory-efficient exact attention with {IO}-awareness.
\newblock In \emph{Advances in Neural Information Processing Systems (NeurIPS)}, 2022.

\bibitem[DeTone et~al.(2017)DeTone, Malisiewicz, and Rabinovich]{DeTone2017SuperPointSI}
Daniel DeTone, Tomasz Malisiewicz, and Andrew Rabinovich.
\newblock Superpoint: Self-supervised interest point detection and description.
\newblock \emph{2018 IEEE/CVF Conference on Computer Vision and Pattern Recognition Workshops (CVPRW)}, pages 337--33712, 2017.

\bibitem[Doersch et~al.(2022)Doersch, Gupta, Markeeva, Recasens, Smaira, Aytar, Carreira, Zisserman, and Yang]{doersch2022tapvid}
Carl Doersch, Ankush Gupta, Larisa Markeeva, Adrià Recasens, Lucas Smaira, Yusuf Aytar, João Carreira, Andrew Zisserman, and Yi Yang.
\newblock Tap-vid: A benchmark for tracking any point in a video, 2022.

\bibitem[Dosovitskiy et~al.(2021)Dosovitskiy, Beyer, Kolesnikov, Weissenborn, Zhai, Unterthiner, Dehghani, Minderer, Heigold, Gelly, Uszkoreit, and Houlsby]{dosovitskiy2021an}
Alexey Dosovitskiy, Lucas Beyer, Alexander Kolesnikov, Dirk Weissenborn, Xiaohua Zhai, Thomas Unterthiner, Mostafa Dehghani, Matthias Minderer, Georg Heigold, Sylvain Gelly, Jakob Uszkoreit, and Neil Houlsby.
\newblock An image is worth 16x16 words: Transformers for image recognition at scale.
\newblock In \emph{International Conference on Learning Representations}, 2021.

\bibitem[Dubey et~al.(2024)Dubey, Jauhri, Pandey, Kadian, Al-Dahle, Letman, Mathur, Schelten, Yang, Fan, et~al.]{dubey2024llama3herdmodels}
Abhimanyu Dubey, Abhinav Jauhri, Abhinav Pandey, Abhishek Kadian, Ahmad Al-Dahle, Aiesha Letman, Akhil Mathur, Alan Schelten, Amy Yang, Angela Fan, et~al.
\newblock The llama 3 herd of models, 2024.

\bibitem[Dusmanu et~al.(2019)Dusmanu, Rocco, Pajdla, Pollefeys, Sivic, Torii, and Sattler]{Dusmanu2019D2NetAT}
Mihai Dusmanu, Ignacio Rocco, Tom{\'a}s Pajdla, Marc Pollefeys, Josef Sivic, Akihiko Torii, and Torsten Sattler.
\newblock D2-net: A trainable cnn for joint description and detection of local features.
\newblock \emph{2019 IEEE/CVF Conference on Computer Vision and Pattern Recognition (CVPR)}, pages 8084--8093, 2019.

\bibitem[Eftekhar et~al.(2021)Eftekhar, Sax, Malik, and Zamir]{eftekhar2021omnidata}
Ainaz Eftekhar, Alexander Sax, Jitendra Malik, and Amir Zamir.
\newblock Omnidata: A scalable pipeline for making multi-task mid-level vision datasets from 3d scans.
\newblock In \emph{Proceedings of the IEEE/CVF International Conference on Computer Vision}, pages 10786--10796, 2021.

\bibitem[{FairScale authors}(2021)]{FairScale2021}
{FairScale authors}.
\newblock Fairscale: A general purpose modular pytorch library for high performance and large scale training.
\newblock \url{https://github.com/facebookresearch/fairscale}, 2021.

\bibitem[Fan et~al.(2024)Fan, Cong, Wen, Wang, Zhang, Ding, Xu, Ivanovic, Pavone, Pavlakos, Wang, and Wang]{fan2024instantsplat}
Zhiwen Fan, Wenyan Cong, Kairun Wen, Kevin Wang, Jian Zhang, Xinghao Ding, Danfei Xu, Boris Ivanovic, Marco Pavone, Georgios Pavlakos, Zhangyang Wang, and Yue Wang.
\newblock Instantsplat: Unbounded sparse-view pose-free gaussian splatting in 40 seconds, 2024.

\bibitem[Galliani et~al.(2015)Galliani, Lasinger, and Schindler]{galliani2015massively}
Silvano Galliani, Katrin Lasinger, and Konrad Schindler.
\newblock Massively parallel multiview stereopsis by surface normal diffusion.
\newblock In \emph{Proceedings of the IEEE international conference on computer vision}, pages 873--881, 2015.

\bibitem[Gleize et~al.(2023)Gleize, Wang, and Feiszli]{Gleize_2023_ICCV_silk}
Pierre Gleize, Weiyao Wang, and Matt Feiszli.
\newblock Silk: Simple learned keypoints.
\newblock In \emph{Proceedings of the IEEE/CVF International Conference on Computer Vision (ICCV)}, pages 10932--10942, 2023.

\bibitem[Grauman et~al.(2024)Grauman, Westbury, Torresani, Kitani, Malik, Afouras, Ashutosh, Baiyya, Bansal, Boote, et~al.]{grauman2024egoexo4d}
Kristen Grauman, Andrew Westbury, Lorenzo Torresani, Kris Kitani, Jitendra Malik, Triantafyllos Afouras, Kumar Ashutosh, Vijay Baiyya, Siddhant Bansal, Bikram Boote, et~al.
\newblock Ego-exo4d: Understanding skilled human activity from first-and third-person perspectives.
\newblock In \emph{Proceedings of the IEEE conference on computer vision and pattern recognition}, 2024.

\bibitem[Hartley and Zisserman(2003)]{HartleyZisserman2003MVG}
Richard Hartley and Andrew Zisserman.
\newblock \emph{Multiple View Geometry in Computer Vision}.
\newblock Cambridge University Press, New York, NY, USA, 2 edition, 2003.

\bibitem[Huang et~al.(2019)Huang, Cheng, Bapna, Firat, Chen, Chen, Lee, Ngiam, Le, Wu, and Chen]{gpipe}
Yanping Huang, Youlong Cheng, Ankur Bapna, Orhan Firat, Dehao Chen, Mia Chen, HyoukJoong Lee, Jiquan Ngiam, Quoc~V Le, Yonghui Wu, and zhifeng Chen.
\newblock Gpipe: Efficient training of giant neural networks using pipeline parallelism.
\newblock In \emph{Advances in Neural Information Processing Systems}. Curran Associates, Inc., 2019.

\bibitem[Jiang et~al.(2024)Jiang, Sablayrolles, Roux, Mensch, Savary, Bamford, Chaplot, de~las Casas, Hanna, Bressand, Lengyel, Bour, Lample, Lavaud, Saulnier, Lachaux, Stock, Subramanian, Yang, Antoniak, Scao, Gervet, Lavril, Wang, Lacroix, and Sayed]{jiang2024mixtralexperts}
Albert~Q. Jiang, Alexandre Sablayrolles, Antoine Roux, Arthur Mensch, Blanche Savary, Chris Bamford, Devendra~Singh Chaplot, Diego de~las Casas, Emma~Bou Hanna, Florian Bressand, Gianna Lengyel, Guillaume Bour, Guillaume Lample, Lélio~Renard Lavaud, Lucile Saulnier, Marie-Anne Lachaux, Pierre Stock, Sandeep Subramanian, Sophia Yang, Szymon Antoniak, Teven~Le Scao, Théophile Gervet, Thibaut Lavril, Thomas Wang, Timothée Lacroix, and William~El Sayed.
\newblock Mixtral of experts, 2024.

\bibitem[Leroy et~al.(2024{\natexlab{a}})Leroy, Cabon, and Revaud]{leroy2024grounding}
Vincent Leroy, Yohann Cabon, and J{\'e}r{\^o}me Revaud.
\newblock Grounding image matching in 3d with mast3r.
\newblock \emph{arXiv preprint arXiv:2406.09756}, 2024{\natexlab{a}}.

\bibitem[Leroy et~al.(2024{\natexlab{b}})Leroy, Cabon, and Revaud]{mast3r_arxiv24}
Vincent Leroy, Yohann Cabon, and Jerome Revaud.
\newblock Grounding image matching in 3d with mast3r, 2024{\natexlab{b}}.

\bibitem[Li et~al.(2020)Li, Zhao, Varma, Salpekar, Noordhuis, Li, Paszke, Smith, Vaughan, Damania, and Chintala]{pytorch-distributed}
Shen Li, Yanli Zhao, Rohan Varma, Omkar Salpekar, Pieter Noordhuis, Teng Li, Adam Paszke, Jeff Smith, Brian Vaughan, Pritam Damania, and Soumith Chintala.
\newblock Pytorch distributed: Experiences on accelerating data parallel training.
\newblock \emph{CoRR}, abs/2006.15704, 2020.

\bibitem[Li and Snavely(2018)]{megadepth}
Zhengqi Li and Noah Snavely.
\newblock Megadepth: Learning single-view depth prediction from internet photos.
\newblock In \emph{Computer Vision and Pattern Recognition (CVPR)}, 2018.

\bibitem[Lin et~al.(2023)Lin, Zhang, Ramanan, and Tulsiani]{lin2023relpose++}
Amy Lin, Jason~Y Zhang, Deva Ramanan, and Shubham Tulsiani.
\newblock Relpose++: Recovering 6d poses from sparse-view observations.
\newblock \emph{arXiv preprint arXiv:2305.04926}, 2023.

\bibitem[Lindenberger et~al.(2021)Lindenberger, Sarlin, Larsson, and Pollefeys]{Lindenberger2021PixelPerfectSW}
Philipp Lindenberger, Paul-Edouard Sarlin, Viktor Larsson, and Marc Pollefeys.
\newblock Pixel-perfect structure-from-motion with featuremetric refinement.
\newblock \emph{2021 IEEE/CVF International Conference on Computer Vision (ICCV)}, pages 5967--5977, 2021.

\bibitem[Loshchilov and Hutter(2017)]{adam-w}
Ilya Loshchilov and Frank Hutter.
\newblock Fixing weight decay regularization in adam.
\newblock \emph{CoRR}, abs/1711.05101, 2017.

\bibitem[Mur-Artal and Tard\'os(2017)]{murORB2}
Ra\'ul Mur-Artal and Juan~D. Tard\'os.
\newblock {ORB-SLAM2}: an open-source {SLAM} system for monocular, stereo and {RGB-D} cameras.
\newblock \emph{IEEE Transactions on Robotics}, 33\penalty0 (5):\penalty0 1255--1262, 2017.

\bibitem[Mur-Artal et~al.(2015)Mur-Artal, Montiel, and Tardos]{mur2015orb}
Raul Mur-Artal, Jose Maria~Martinez Montiel, and Juan~D Tardos.
\newblock Orb-slam: a versatile and accurate monocular slam system.
\newblock \emph{IEEE transactions on robotics}, 31\penalty0 (5):\penalty0 1147--1163, 2015.

\bibitem[Narayanan et~al.(2021)Narayanan, Shoeybi, Casper, LeGresley, Patwary, Korthikanti, Vainbrand, Kashinkunti, Bernauer, Catanzaro, Phanishayee, and Zaharia]{megatron-lm2}
Deepak Narayanan, Mohammad Shoeybi, Jared Casper, Patrick LeGresley, Mostofa Patwary, Vijay Korthikanti, Dmitri Vainbrand, Prethvi Kashinkunti, Julie Bernauer, Bryan Catanzaro, Amar Phanishayee, and Matei Zaharia.
\newblock Efficient large-scale language model training on {GPU} clusters.
\newblock \emph{CoRR}, abs/2104.04473, 2021.

\bibitem[Oquab et~al.(2023)Oquab, Darcet, Moutakanni, Vo, Szafraniec, Khalidov, Fernandez, Haziza, Massa, El-Nouby, Howes, Huang, Xu, Sharma, Li, Galuba, Rabbat, Assran, Ballas, Synnaeve, Misra, Jegou, Mairal, Labatut, Joulin, and Bojanowski]{oquab2023dinov2}
Maxime Oquab, Timothée Darcet, Theo Moutakanni, Huy~V. Vo, Marc Szafraniec, Vasil Khalidov, Pierre Fernandez, Daniel Haziza, Francisco Massa, Alaaeldin El-Nouby, Russell Howes, Po-Yao Huang, Hu Xu, Vasu Sharma, Shang-Wen Li, Wojciech Galuba, Mike Rabbat, Mido Assran, Nicolas Ballas, Gabriel Synnaeve, Ishan Misra, Herve Jegou, Julien Mairal, Patrick Labatut, Armand Joulin, and Piotr Bojanowski.
\newblock Dinov2: Learning robust visual features without supervision, 2023.

\bibitem[Raistrick et~al.(2023)Raistrick, Lipson, Ma, Mei, Wang, Zuo, Kayan, Wen, Han, Wang, Newell, Law, Goyal, Yang, and Deng]{infinigen}
Alexander Raistrick, Lahav Lipson, Zeyu Ma, Lingjie Mei, Mingzhe Wang, Yiming Zuo, Karhan Kayan, Hongyu Wen, Beining Han, Yihan Wang, Alejandro Newell, Hei Law, Ankit Goyal, Kaiyu Yang, and Jia Deng.
\newblock Infinite photorealistic worlds using procedural generation.
\newblock In \emph{Proceedings of the IEEE/CVF Conference on Computer Vision and Pattern Recognition}, pages 12630--12641, 2023.

\bibitem[Rajbhandari et~al.(2019)Rajbhandari, Rasley, Ruwase, and He]{deepspeed-zero}
Samyam Rajbhandari, Jeff Rasley, Olatunji Ruwase, and Yuxiong He.
\newblock Zero: Memory optimization towards training {A} trillion parameter models.
\newblock \emph{CoRR}, abs/1910.02054, 2019.

\bibitem[Ranftl et~al.(2020)Ranftl, Lasinger, Hafner, Schindler, and Koltun]{Ranftl2020midas}
Ren\'{e} Ranftl, Katrin Lasinger, David Hafner, Konrad Schindler, and Vladlen Koltun.
\newblock Towards robust monocular depth estimation: Mixing datasets for zero-shot cross-dataset transfer.
\newblock In \emph{IEEE Transactions on Pattern Analysis and Machine Intelligence (TPAMI)}, 2020.

\bibitem[Ranftl et~al.(2021)Ranftl, Bochkovskiy, and Koltun]{ranftl2021dpt}
Ren\'{e} Ranftl, Alexey Bochkovskiy, and Vladlen Koltun.
\newblock Vision transformers for dense prediction.
\newblock \emph{arXiv preprint arXiv:2103.13413}, 2021.

\bibitem[Rasley et~al.(2020)Rasley, Rajbhandari, Ruwase, and He]{deepspeed}
Jeff Rasley, Samyam Rajbhandari, Olatunji Ruwase, and Yuxiong He.
\newblock Deepspeed: System optimizations enable training deep learning models with over 100 billion parameters.
\newblock In \emph{Proceedings of the 26th ACM SIGKDD International Conference on Knowledge Discovery \& Data Mining}, pages 3505--3506, 2020.

\bibitem[Reizenstein et~al.(2021)Reizenstein, Shapovalov, Henzler, Sbordone, Labatut, and Novotny]{reizenstein2021common}
Jeremy Reizenstein, Roman Shapovalov, Philipp Henzler, Luca Sbordone, Patrick Labatut, and David Novotny.
\newblock Common objects in 3d: Large-scale learning and evaluation of real-life 3d category reconstruction.
\newblock In \emph{Proceedings of the IEEE/CVF international conference on computer vision}, pages 10901--10911, 2021.

\bibitem[Roberts et~al.(2021)Roberts, Ramapuram, Ranjan, Kumar, Bautista, Paczan, Webb, and Susskind]{hypersim}
Mike Roberts, Jason Ramapuram, Anurag Ranjan, Atulit Kumar, Miguel~Angel Bautista, Nathan Paczan, Russ Webb, and Joshua~M. Susskind.
\newblock {Hypersim}: {A} photorealistic synthetic dataset for holistic indoor scene understanding.
\newblock In \emph{International Conference on Computer Vision (ICCV) 2021}, 2021.

\bibitem[Sarlin et~al.(2019)Sarlin, DeTone, Malisiewicz, and Rabinovich]{Sarlin2019SuperGlueLF}
Paul-Edouard Sarlin, Daniel DeTone, Tomasz Malisiewicz, and Andrew Rabinovich.
\newblock Superglue: Learning feature matching with graph neural networks.
\newblock \emph{2020 IEEE/CVF Conference on Computer Vision and Pattern Recognition (CVPR)}, pages 4937--4946, 2019.

\bibitem[Sarlin et~al.(2020)Sarlin, DeTone, Malisiewicz, and Rabinovich]{sarlin20superglue}
Paul-Edouard Sarlin, Daniel DeTone, Tomasz Malisiewicz, and Andrew Rabinovich.
\newblock {SuperGlue}: Learning feature matching with graph neural networks.
\newblock In \emph{CVPR}, 2020.

\bibitem[Savva et~al.(2019)Savva, Kadian, Maksymets, Zhao, Wijmans, Jain, Straub, Liu, Koltun, Malik, Parikh, and Batra]{habitat}
Manolis Savva, Abhishek Kadian, Oleksandr Maksymets, Yili Zhao, Erik Wijmans, Bhavana Jain, Julian Straub, Jia Liu, Vladlen Koltun, Jitendra Malik, Devi Parikh, and Dhruv Batra.
\newblock Habitat: {A} platform for embodied {AI} research.
\newblock \emph{CoRR}, abs/1904.01201, 2019.

\bibitem[Sch\"{o}nberger and Frahm(2016)]{schoenberger2016sfmColmap}
Johannes~Lutz Sch\"{o}nberger and Jan-Michael Frahm.
\newblock Structure-from-motion revisited.
\newblock In \emph{Conference on Computer Vision and Pattern Recognition (CVPR)}, 2016.

\bibitem[Shazeer et~al.(2018)Shazeer, Cheng, Parmar, Tran, Vaswani, Koanantakool, Hawkins, Lee, Hong, Young, Sepassi, and Hechtman]{mesh-tf}
Noam Shazeer, Youlong Cheng, Niki Parmar, Dustin Tran, Ashish Vaswani, Penporn Koanantakool, Peter Hawkins, HyoukJoong Lee, Mingsheng Hong, Cliff Young, Ryan Sepassi, and Blake~A. Hechtman.
\newblock Mesh-tensorflow: Deep learning for supercomputers.
\newblock \emph{CoRR}, abs/1811.02084, 2018.

\bibitem[Shoeybi et~al.(2019)Shoeybi, Patwary, Puri, LeGresley, Casper, and Catanzaro]{megatron-lm}
Mohammad Shoeybi, Mostofa Patwary, Raul Puri, Patrick LeGresley, Jared Casper, and Bryan Catanzaro.
\newblock Megatron-lm: Training multi-billion parameter language models using model parallelism.
\newblock \emph{CoRR}, abs/1909.08053, 2019.

\bibitem[Shotton et~al.(2013)Shotton, Glocker, Zach, Izadi, Criminisi, and Fitzgibbon]{shotton2013scene}
Jamie Shotton, Ben Glocker, Christopher Zach, Shahram Izadi, Antonio Criminisi, and Andrew Fitzgibbon.
\newblock Scene coordinate regression forests for camera relocalization in rgb-d images.
\newblock In \emph{Proceedings of the IEEE conference on computer vision and pattern recognition}, pages 2930--2937, 2013.

\bibitem[Smith et~al.(2024)Smith, Charatan, Tewari, and Sitzmann]{Smith2024FlowMapHC}
Cameron Smith, David Charatan, Ayush~Kumar Tewari, and Vincent Sitzmann.
\newblock Flowmap: High-quality camera poses, intrinsics, and depth via gradient descent.
\newblock \emph{ArXiv}, abs/2404.15259, 2024.

\bibitem[Tancik et~al.(2020)Tancik, Srinivasan, Mildenhall, Fridovich-Keil, Raghavan, Singhal, Ramamoorthi, Barron, and Ng]{tancik2020fourier}
Matthew Tancik, Pratul~P. Srinivasan, Ben Mildenhall, Sara Fridovich-Keil, Nithin Raghavan, Utkarsh Singhal, Ravi Ramamoorthi, Jonathan~T. Barron, and Ren Ng.
\newblock Fourier features let networks learn high frequency functions in low dimensional domains.
\newblock In \emph{Advances in Neural Information Processing Systems}, pages 7537--7547, 2020.

\bibitem[Tang et~al.(2024)Tang, Wang, and Feiszli]{Tang2024ADen}
Hao Tang, Weiyao Wang, and Matt Feiszli.
\newblock Aden: Adaptive density representations for sparse-view camera pose estimation.
\newblock \emph{arXiv preprint arXiv:2408.09042}, 2024.

\bibitem[Teed and Deng(2021{\natexlab{a}})]{Teed2021DROIDSLAMDV}
Zachary Teed and Jia Deng.
\newblock Droid-slam: Deep visual slam for monocular, stereo, and rgb-d cameras.
\newblock In \emph{Neural Information Processing Systems}, 2021{\natexlab{a}}.

\bibitem[Teed and Deng(2021{\natexlab{b}})]{teed2021droid}
Zachary Teed and Jia Deng.
\newblock {DROID-SLAM: Deep Visual SLAM for Monocular, Stereo, and RGB-D Cameras}.
\newblock \emph{Advances in neural information processing systems}, 2021{\natexlab{b}}.

\bibitem[Thrun(2002)]{thrun2002probabilistic}
Sebastian Thrun.
\newblock Probabilistic robotics.
\newblock \emph{Communications of the ACM}, 45\penalty0 (3):\penalty0 52--57, 2002.

\bibitem[Touvron et~al.(2023)Touvron, Martin, Stone, Albert, Almahairi, Babaei, Bashlykov, Batra, Bhargava, Bhosale, et~al.]{touvron2023llama2openfoundation}
Hugo Touvron, Louis Martin, Kevin Stone, Peter Albert, Amjad Almahairi, Yasmine Babaei, Nikolay Bashlykov, Soumya Batra, Prajjwal Bhargava, Shruti Bhosale, et~al.
\newblock Llama 2: Open foundation and fine-tuned chat models, 2023.

\bibitem[Tschernezki et~al.(2023)Tschernezki, Darkhalil, Zhu, Fouhey, Larina, Larlus, Damen, and Vedaldi]{EPICFields2023}
Vadim Tschernezki, Ahmad Darkhalil, Zhifan Zhu, David Fouhey, Iro Larina, Diane Larlus, Dima Damen, and Andrea Vedaldi.
\newblock {EPIC Fields}: {M}arrying {3D} {G}eometry and {V}ideo {U}nderstanding.
\newblock In \emph{Proceedings of the Neural Information Processing Systems (NeurIPS)}, 2023.

\bibitem[Vaswani et~al.(2017)Vaswani, Shazeer, Parmar, Uszkoreit, Jones, Gomez, Łukasz Kaiser, and Polosukhin]{vaswani2017attention}
Ashish Vaswani, Noam Shazeer, Niki Parmar, Jakob Uszkoreit, Llion Jones, Aidan~N. Gomez, Łukasz Kaiser, and Illia Polosukhin.
\newblock Attention is all you need.
\newblock \emph{Advances in Neural Information Processing Systems}, 2017.

\bibitem[Wang and Agapito(2024)]{wang2024spann3r}
Hengyi Wang and Lourdes Agapito.
\newblock 3d reconstruction with spatial memory.
\newblock \emph{arXiv preprint arXiv:2408.16061}, 2024.

\bibitem[Wang et~al.(2023{\natexlab{a}})Wang, Karaev, Rupprecht, and Novotny]{Wang2023VGGSfMVG}
Jianyuan Wang, Nikita Karaev, Christian Rupprecht, and David Novotny.
\newblock Vggsfm: Visual geometry grounded deep structure from motion.
\newblock \emph{2024 IEEE/CVF Conference on Computer Vision and Pattern Recognition (CVPR)}, pages 21686--21697, 2023{\natexlab{a}}.

\bibitem[Wang et~al.(2023{\natexlab{b}})Wang, Rupprecht, and Novotn{\'y}]{Wang2023PoseDiffusionSP}
Jianyuan Wang, C. Rupprecht, and David Novotn{\'y}.
\newblock Posediffusion: Solving pose estimation via diffusion-aided bundle adjustment.
\newblock \emph{2023 IEEE/CVF International Conference on Computer Vision (ICCV)}, pages 9739--9749, 2023{\natexlab{b}}.

\bibitem[Wang et~al.(2023{\natexlab{c}})Wang, Leroy, Cabon, Chidlovskii, and Revaud]{Wang2023DUSt3RG3}
Shuzhe Wang, Vincent Leroy, Yohann Cabon, Boris Chidlovskii, and J{\'e}r{\^o}me Revaud.
\newblock Dust3r: Geometric 3d vision made easy.
\newblock \emph{2024 IEEE/CVF Conference on Computer Vision and Pattern Recognition (CVPR)}, pages 20697--20709, 2023{\natexlab{c}}.

\bibitem[Wang et~al.(2024)Wang, Leroy, Cabon, Chidlovskii, and Revaud]{dust3r_cvpr24}
Shuzhe Wang, Vincent Leroy, Yohann Cabon, Boris Chidlovskii, and Jerome Revaud.
\newblock Dust3r: Geometric 3d vision made easy.
\newblock In \emph{CVPR}, 2024.

\bibitem[Wang et~al.(2020)Wang, Zhu, Wang, Hu, Qiu, Wang, Hu, Kapoor, and Scherer]{tartanair2020iros}
Wenshan Wang, Delong Zhu, Xiangwei Wang, Yaoyu Hu, Yuheng Qiu, Chen Wang, Yafei Hu, Ashish Kapoor, and Sebastian Scherer.
\newblock Tartanair: A dataset to push the limits of visual slam.
\newblock In \emph{2020 IEEE/RSJ International Conference on Intelligent Robots and Systems (IROS)}, 2020.

\bibitem[Weinzaepfel et~al.(2022)Weinzaepfel, Leroy, Lucas, Br{\'e}gier, Cabon, Arora, Antsfeld, Chidlovskii, Csurka, and Revaud]{weinzaepfel2022croco}
Philippe Weinzaepfel, Vincent Leroy, Thomas Lucas, Romain Br{\'e}gier, Yohann Cabon, Vaibhav Arora, Leonid Antsfeld, Boris Chidlovskii, Gabriela Csurka, and J{\'e}r{\^o}me Revaud.
\newblock Croco: Self-supervised pre-training for 3d vision tasks by cross-view completion.
\newblock \emph{Advances in Neural Information Processing Systems}, 35:\penalty0 3502--3516, 2022.

\bibitem[Wolf et~al.(2019)Wolf, Debut, Sanh, Chaumond, Delangue, Moi, Cistac, Rault, Louf, Funtowicz, and Brew]{hf-transformers}
Thomas Wolf, Lysandre Debut, Victor Sanh, Julien Chaumond, Clement Delangue, Anthony Moi, Pierric Cistac, Tim Rault, R{\'{e}}mi Louf, Morgan Funtowicz, and Jamie Brew.
\newblock Huggingface's transformers: State-of-the-art natural language processing.
\newblock \emph{CoRR}, abs/1910.03771, 2019.

\bibitem[Xu et~al.(2023)Xu, Yin, Chen, Shen, Cheng, and Zhao]{xu2023frozenrecon}
Guangkai Xu, Wei Yin, Hao Chen, Chunhua Shen, Kai Cheng, and Feng Zhao.
\newblock Frozenrecon: Pose-free 3d scene reconstruction with frozen depth models.
\newblock In \emph{Proceedings of the IEEE/CVF International Conference on Computer Vision}, pages 9310--9320, 2023.

\bibitem[Yang et~al.(2024)Yang, Kang, Huang, Xu, Feng, and Zhao]{depthanything2_2024}
Lihe Yang, Bingyi Kang, Zilong Huang, Xiaogang Xu, Jiashi Feng, and Hengshuang Zhao.
\newblock Depth anything: Unleashing the power of large-scale unlabeled data.
\newblock In \emph{CVPR}, 2024.

\bibitem[Yao et~al.(2020)Yao, Luo, Li, Zhang, Ren, Zhou, Fang, and Quan]{yao2020blendedmvs}
Yao Yao, Zixin Luo, Shiwei Li, Jingyang Zhang, Yufan Ren, Lei Zhou, Tian Fang, and Long Quan.
\newblock Blendedmvs: A large-scale dataset for generalized multi-view stereo networks.
\newblock \emph{Computer Vision and Pattern Recognition (CVPR)}, 2020.

\bibitem[Yeshwanth et~al.(2023)Yeshwanth, Liu, Nie{\ss}ner, and Dai]{yeshwanth2023scannet++}
Chandan Yeshwanth, Yueh-Cheng Liu, Matthias Nie{\ss}ner, and Angela Dai.
\newblock Scannet++: A high-fidelity dataset of 3d indoor scenes.
\newblock In \emph{Proceedings of the IEEE/CVF International Conference on Computer Vision}, pages 12--22, 2023.

\bibitem[Yi et~al.(2016)Yi, Trulls, Lepetit, and Fua]{Yi2016LIFTLI}
Kwang~Moo Yi, Eduard Trulls, Vincent Lepetit, and Pascal~V. Fua.
\newblock Lift: Learned invariant feature transform.
\newblock In \emph{European Conference on Computer Vision}, 2016.

\bibitem[Zhang et~al.(2024{\natexlab{a}})Zhang, Herrmann, Hur, Jampani, Darrell, Cole, Sun, and Yang]{zhang2024monst3r}
Junyi Zhang, Charles Herrmann, Junhwa Hur, Varun Jampani, Trevor Darrell, Forrester Cole, Deqing Sun, and Ming-Hsuan Yang.
\newblock Monst3r: A simple approach for estimating geometry in the presence of motion.
\newblock \emph{arXiv preprint arxiv:2410.03825}, 2024{\natexlab{a}}.

\bibitem[Zhang et~al.(2022)Zhang, Ramanan, and Tulsiani]{Zhang2022RelPosePP}
Jason~Y. Zhang, Deva Ramanan, and Shubham Tulsiani.
\newblock Relpose: Predicting probabilistic relative rotation for single objects in the wild.
\newblock \emph{ArXiv}, abs/2208.05963, 2022.

\bibitem[Zhang et~al.(2024{\natexlab{b}})Zhang, Lin, Kumar, Yang, Ramanan, and Tulsiani]{Zhang2024CamerasAR}
Jason~Y. Zhang, Amy Lin, Moneish Kumar, Tzu-Hsuan Yang, Deva Ramanan, and Shubham Tulsiani.
\newblock Cameras as rays: Pose estimation via ray diffusion.
\newblock \emph{ArXiv}, abs/2402.14817, 2024{\natexlab{b}}.

\bibitem[Zhao et~al.(2022)Zhao, Liu, Guo, Wang, and Liu]{Zhao2022ParticleSfMED}
Wang Zhao, Shaohui Liu, Hengkai Guo, Wenping Wang, and Y. Liu.
\newblock Particlesfm: Exploiting dense point trajectories for localizing moving cameras in the wild.
\newblock In \emph{European Conference on Computer Vision}, 2022.

\bibitem[Zheng et~al.(2023)Zheng, Harley, Shen, Wetzstein, and Guibas]{zheng2023pointodyssey}
Yang Zheng, Adam~W Harley, Bokui Shen, Gordon Wetzstein, and Leonidas~J Guibas.
\newblock Pointodyssey: A large-scale synthetic dataset for long-term point tracking.
\newblock In \emph{Proceedings of the IEEE/CVF International Conference on Computer Vision}, pages 19855--19865, 2023.

\bibitem[Zhou et~al.(2018)Zhou, Tucker, Flynn, Fyffe, and Snavely]{realestate10k}
Tinghui Zhou, Richard Tucker, John Flynn, Graham Fyffe, and Noah Snavely.
\newblock Stereo magnification: learning view synthesis using multiplane images.
\newblock \emph{ACM Trans. Graph.}, 37\penalty0 (4), 2018.

\bibitem[Zhu et~al.()Zhu, Yang, Wang, Song, Wu, Wei, and Li]{zhupose}
Dawei Zhu, Nan Yang, Liang Wang, Yifan Song, Wenhao Wu, Furu Wei, and Sujian Li.
\newblock Pose: Efficient context window extension of llms via positional skip-wise training.
\newblock In \emph{The Twelfth International Conference on Learning Representations}.

\end{thebibliography}
}

\clearpage
\appendix
\maketitlesupplementary

\section{Model Scaling Effect}
\label{sec:scaling_model}
\begin{figure*}
    \centering
    \includegraphics[width=\textwidth]{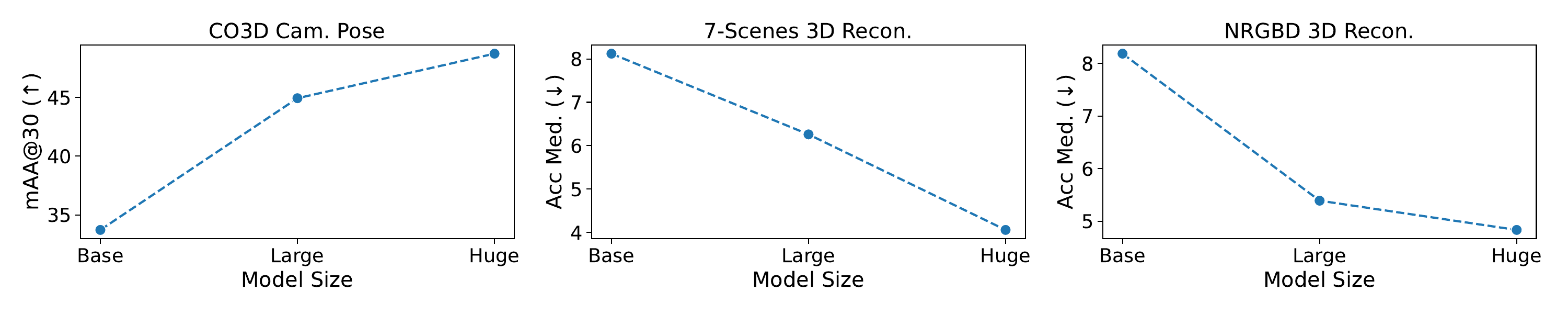}
    \caption{\textbf{Model scaling effect.} Increasing the size of the Fusion Transformer leads to better camera pose estimation ($\uparrow$) and 3D reconstruction ($\downarrow$). All models are trained for 60k steps (equivalent to 60 epochs; the main paper uses 100 epochs).}
    \label{fig:model_scaling}
\end{figure*}

We investigate the effect of scaling model size by trying three model sizes for the Fusion Transformer: ViT-base, ViT-large, and ViT-huge, according to the settings in the original ViT paper \cite{dosovitskiy2021an}. The results are shown in Figure \ref{fig:model_scaling}. This experiment demonstrates that larger model size continually benefits 3D tasks including camera pose estimations and 3D reconstruction. Note that the Fusion Transformer size used in the main text for all experiments is a ViT-base.

\section{Data Scaling Effect}
\label{sec:scaling_data}
\begin{figure*}
    \centering
    \includegraphics[width=\textwidth]{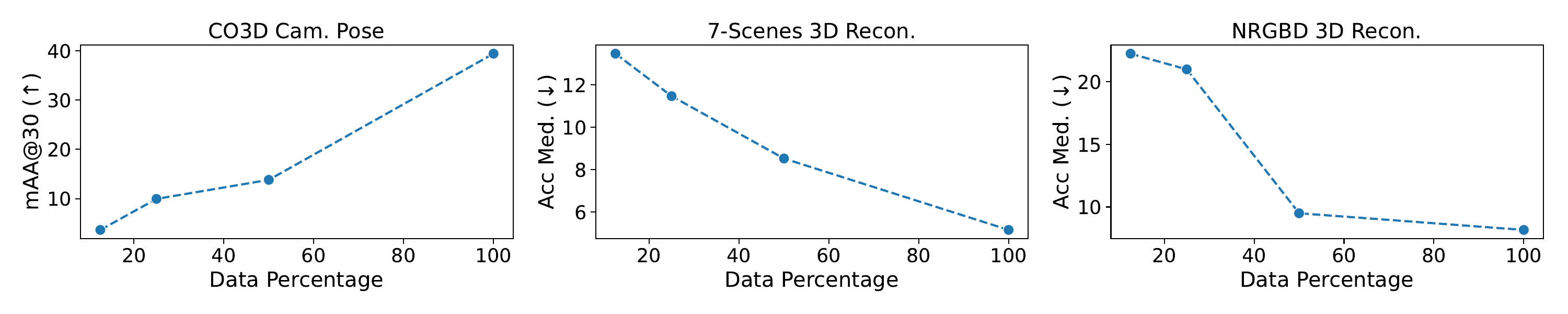}
    \caption{\textbf{Data scaling effect.} More training data leads to better camera pose estimation ($\uparrow$) and 3D reconstruction ($\downarrow$). All models are trained for 60k steps (equivalent to 60 epochs; the main paper uses 100 epochs).}
    \label{fig:data_scaling}
\end{figure*}

We study the effect of scaling the data using 4 different scales of data, $12.5\%$, $25\%$, $50\%$, and $100\%$, to train the model. The results are shown in Figure \ref{fig:data_scaling}. The training settings for all models are kept the same except for how much data they have access to. The results demonstrate that Fast3R continually benefits from more data, suggesting Fast3R could achieve better results in the future given more data.

\section{Gaussian Splatting}
We qualitatively demonstrate the potential of using Fast3R's output for downstream novel view synthesis tasks. A visualization of the Gaussian Splatting generated by adopting the pipeline of InstantSplat \cite{fan2024instantsplat} is shown in Figure \ref{fig:gaussian_vis_co3d}.

\section{Bundle Adjustment (via Gaussian Splatting)}
While not necessary, using bundle adjustment at inference time can also improve Fast3R's performance. We show an example of bundle adjustment using Gaussian Splatting (GS-BA).

\begin{figure*}[h!]
    \centering
    \includegraphics[width=\textwidth]{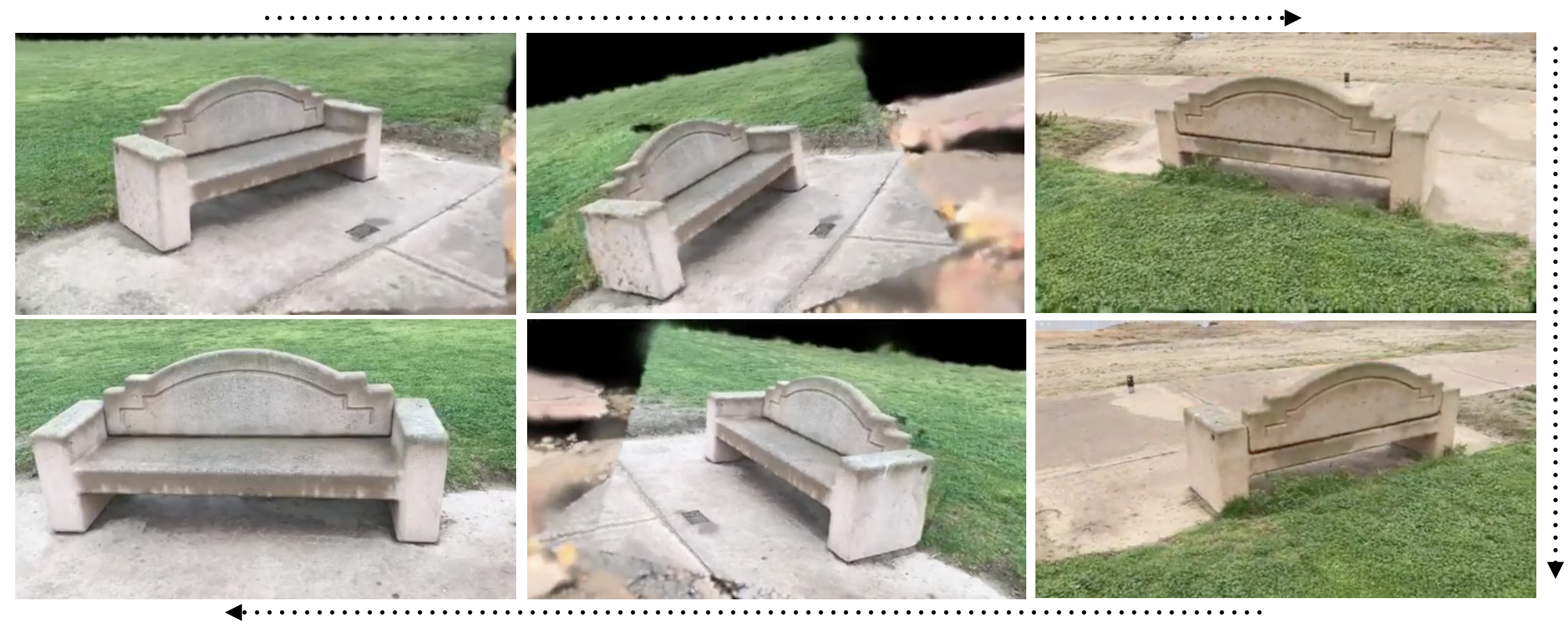} 
    \caption{\textbf{Visualization of Gaussians from unseen poses.} The frames are ordered temporally along the direction of the arrows. The middle frames show poses very different from those used for reconstruction, as is evidenced by the large areas with no Gausisans. The scene is fit from 7 images from CO3D.}
    \label{fig:gaussian_vis_co3d}
\end{figure*}

Specifically, we use InstantSplat \cite{fan2024instantsplat} to optimize a set of Gaussians per scene, using initializations from a point cloud, and update the locations and poses in order to minimize reprojection error. We show an example of the Gaussian reconstruction in Figure~\ref{fig:gaussian_vis_co3d} shows an example reconstruction on CO3D.

We can compare against ground-truth trajectories from COLMAP. We found that GS-BA significantly reduces both the pose and translation error. Table~\ref{tab:gs_ba_quantitative} quantifies this, showing over a 2.5x reduction in translation error and a 4x reduction in rotational error on the ``Family'' scene from Tanks and Temples, which we found to be representative. We show a visualization of the original reconstruction and the poses pre- and post-bundle-adjustment. There are only 8 scenes in the evaluation set in InstantSplat.

\begin{table}[h]
    \centering
    \resizebox{\linewidth}{!}{
    \begin{tabular}{lcc}
        \toprule
        Method & RPE Rotation ($\downarrow$) & RPE Translation  ($\downarrow$) \\
        \midrule
        Fast3R & 27.9 & 7.64\\
        Fast3R w/ GS-BA & \textbf{11.0} & \textbf{1.80} \\
        \bottomrule
    \end{tabular}
    }
    \caption{\textbf{Pose estimation can further improve with Bundle Adjustment.} We show an example on the "Family" scene from Tanks and Temples, using InstantSplat~\cite{fan2024instantsplat}.}
    \label{tab:gs_ba_quantitative}
\end{table}

\begin{figure*}[h!]
    \centering
    \includegraphics[width=\linewidth]{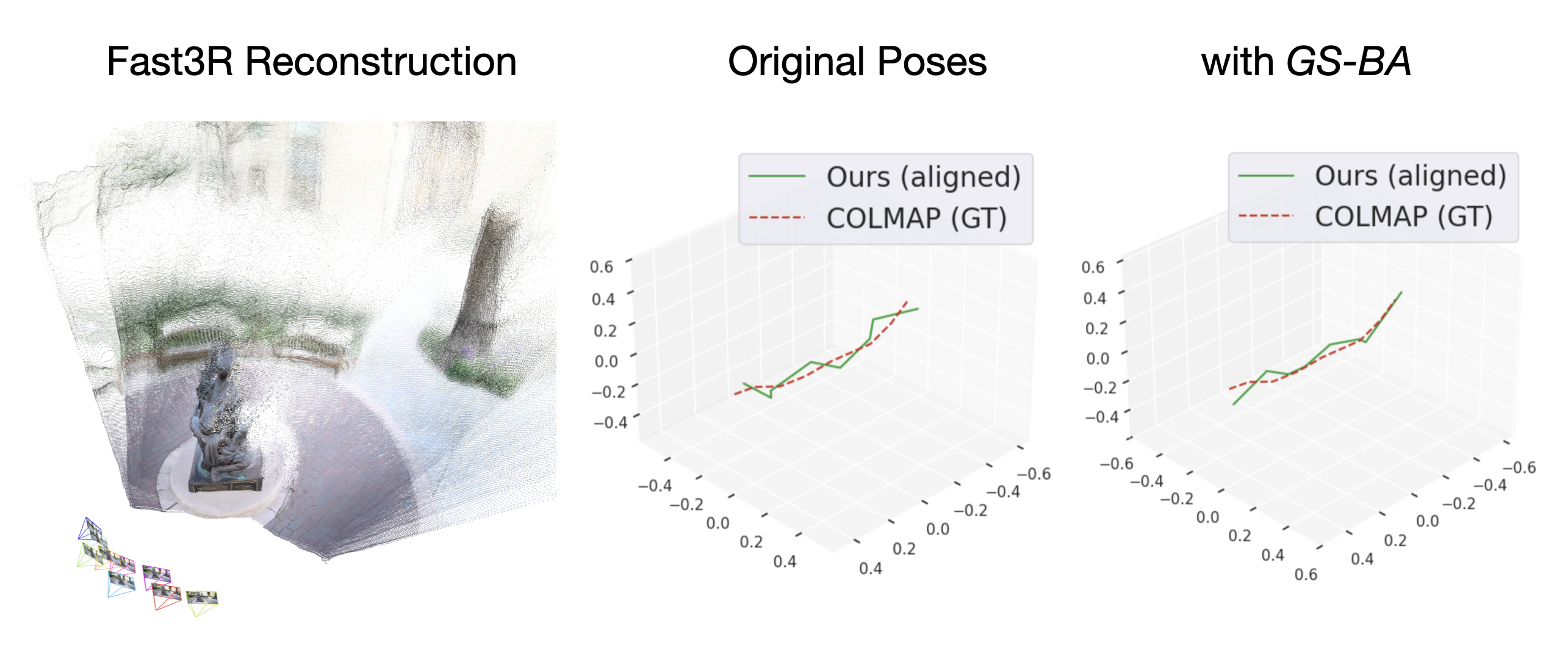} 
    \caption{\textbf{Bundle adjustment further improves pose.} Left: reconstruction from Fast3R. Middle: Original poses pre-GS-BA. Right: Poses after GS-BA.}
    \label{fig:pose_vis_tnt}
\end{figure*}

\begin{table}[t]
\begin{center}
\renewcommand\arraystretch{1.0}
\setlength{\tabcolsep}{3pt} 
\small
\resizebox{\columnwidth}{!}{
\begin{tabular}{llcccccccc}
 \hline
\specialrule{0.5pt}{0.5pt}{0.5pt}
\multicolumn{2}{l}{\multirow{2}{*}{Methods}} & \multicolumn{2}{c}{ScanNet} & \multicolumn{2}{c}{ETH3D} & \multicolumn{2}{c}{DTU} & \multicolumn{2}{c}{T\&T} \\
\cline{3-10}
 & & rel $\downarrow$ & $\tau \uparrow$ & rel $\downarrow$ & $\tau \uparrow$ & rel $\downarrow$ & $\tau \uparrow$ & rel $\downarrow$ & $\tau \uparrow$ \\
\specialrule{1.0pt}{0.5pt}{0.5pt}
\bf{{COLMAP-DENSE}}   & & 38.0 & 22.5 & 89.8 & 23.2 & 20.8 & 69.3 & 25.7 & 76.4 \\
\specialrule{1.5pt}{0.5pt}{0.5pt}
\bf{{DUSt3R 224}}     & & 5.86 & 50.84 & 4.71 & 61.74 & \bf 2.76 & \bf 77.32 & 5.54 & 56.38 \\
\bf{{DUSt3R 512}}     & & \bf 4.93 & \bf 60.20 & \bf 2.91 & \bf 76.91 & 3.52 & 69.33 & \bf 3.17 & \bf 76.68 \\
\specialrule{1.5pt}{0.5pt}{0.5pt}
\bf{{Fast3R}}         & &  6.27 &  50.27 & 4.68  & 62.68 & 3.92 & 62.60 & 4.43 &  63.95 \\

\specialrule{1.5pt}{0.5pt}{0.5pt}
\end{tabular}}
\vspace{-5pt}
\normalsize
\caption{
\textbf{Multi-view depth evaluation.} DUSt3R and Fast3R perform on par, while significantly outperforming COLMAP-DENSE.
}
\label{tab:mvd_selected}
\end{center}
\vspace*{-5pt}
\end{table}

\section{Multi-view Depth Evaluation}
We compare Fast3R (using the local pointmap prediction) with DUSt3R and COLMAP on multi-view depth estimation tasks and show results in Table \ref{tab:mvd_selected}.

\section{More Visualizations}
\begin{figure*}
    \centering
    \includegraphics[width=\textwidth]{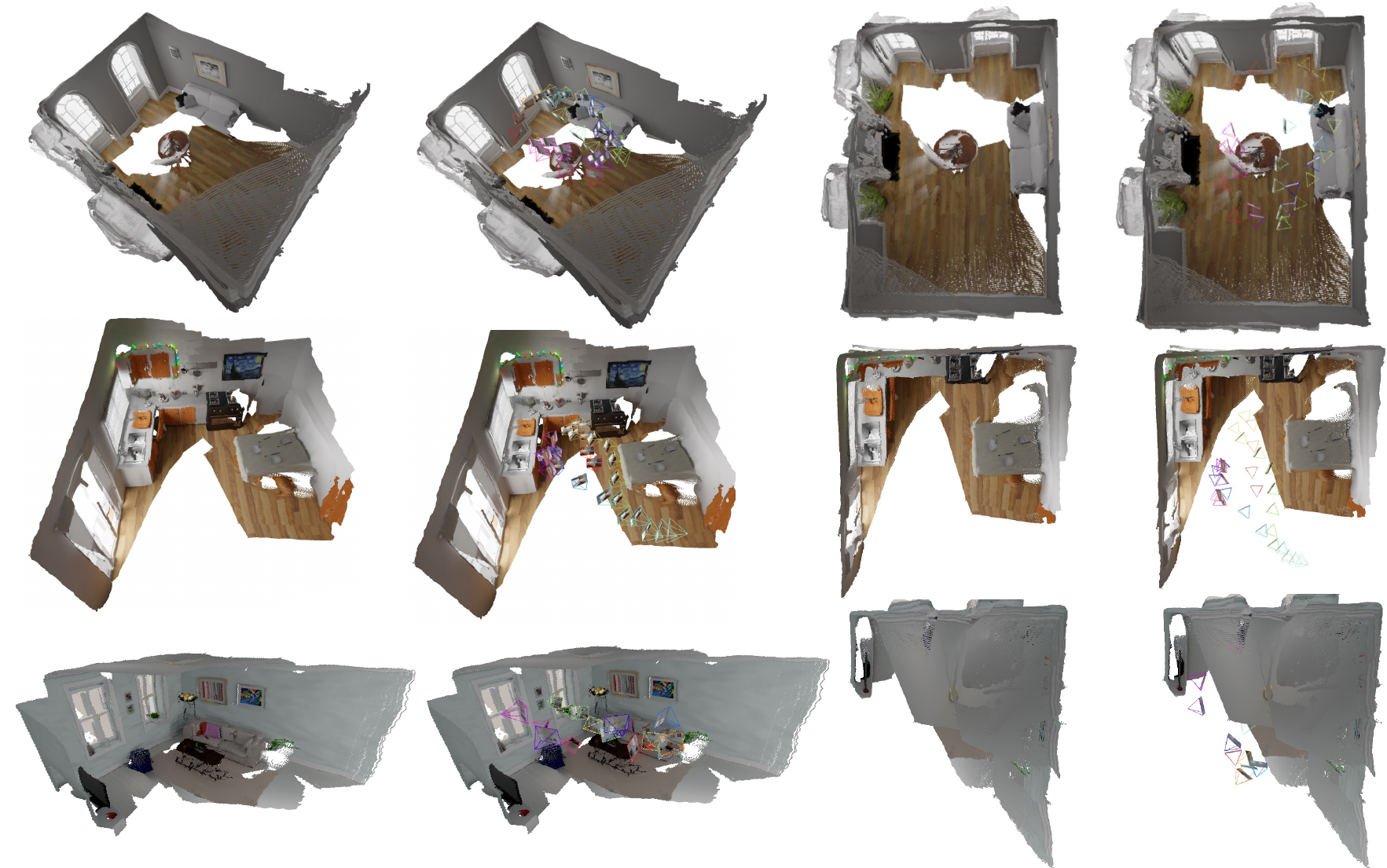}
    \caption{\textbf{Visualizations of results on NRGBD scenes.} Fast3R learns the regularity of indoor rooms (square-like shapes) and demonstrates loop closure capabilities.}
    \label{fig:viz_nrgbd_regularity}
\end{figure*}

We show more visualizations of Fast3R's performance on indoor scenes in Figure \ref{fig:viz_nrgbd_regularity}. Fast3R learns the regularity of indoor rooms (square-like shapes) and demonstrates ``loop closure" capabilities.

\begin{figure*}[t]
    \centering
    \includegraphics[width=\linewidth]{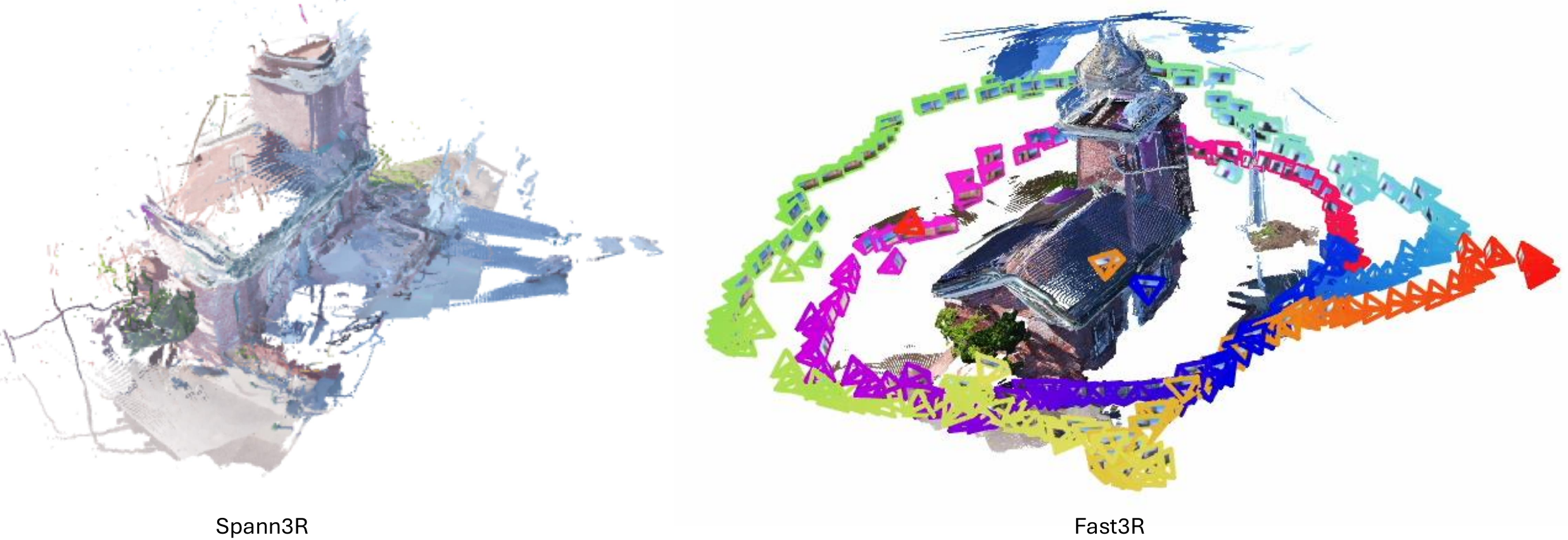}
    \vspace{-18pt}
    \caption{\small{\textbf{Large-scale reconstruction: Spann3R vs. Fast3R on the Lighthouse scene from Tanks \& Temples dataset.}}}
    \label{fig:large_scale}
    \vspace{-15pt}
\end{figure*}

\begin{figure*}[ht]
    \centering
    \includegraphics[width=\linewidth]{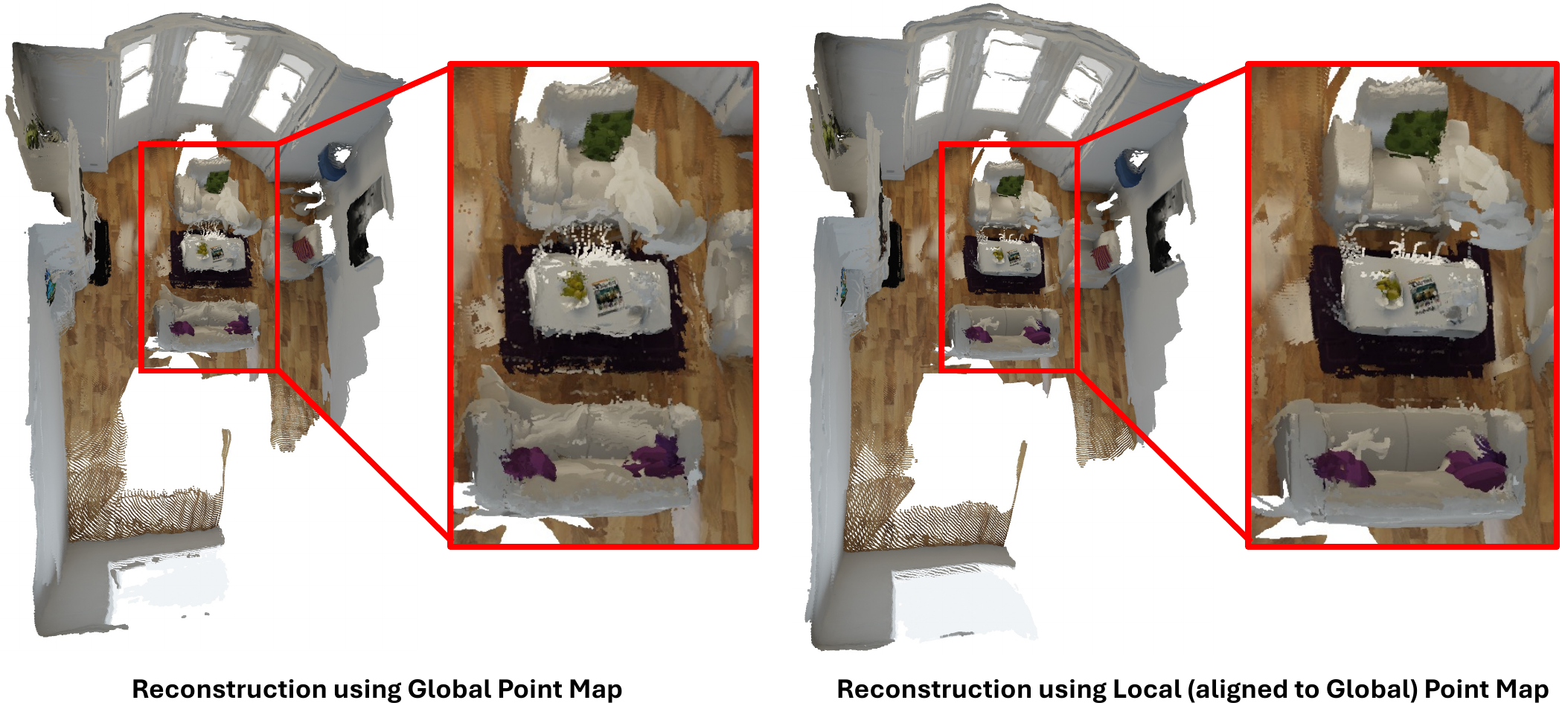}
    \caption{\textbf{Effect of using local vs. global pointmap.} Global point maps provide good anchors for locations of points while local point maps use those anchors (by aligning using ICP on the anchor points to the global point map) to provide more accurate point locations. Best viewed when zoomed in.}
    \label{fig:viz_local_alignment}
\end{figure*}

\begin{figure*}[h!]
    \centering
    \includegraphics[width=\linewidth]{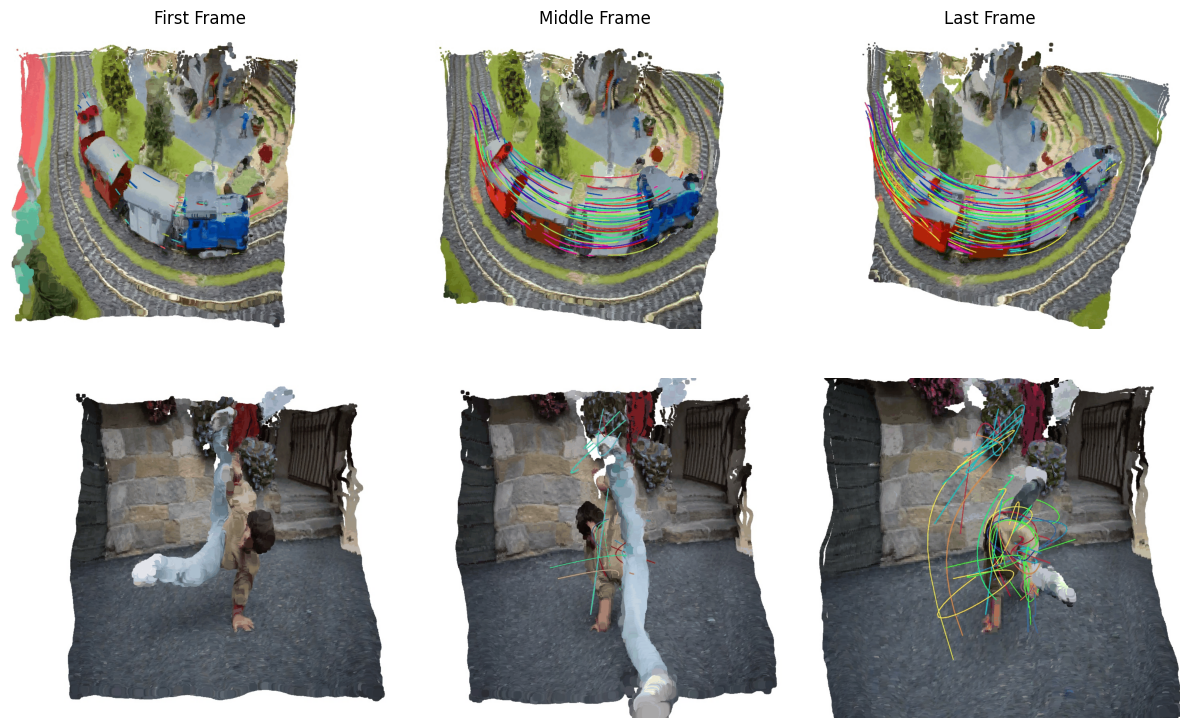}
    \vspace{-4mm}
    \caption{\textbf{Qualitative 4D reconstruction results on unseen dynamic scenes in DAVIS}. Results are obtained with one forward pass. The tracks are visualized using ground-truth track annotations from TAP-Vid-DAVIS \cite{doersch2022tapvid}.}
    \label{fig:4d_reconstruction}
    \vspace{-2mm}
\end{figure*}

\subsection{4D Reconstruction: Qualitative Results}
Because \model{} can handle multiple frames naturally, one may wonder how well \model{} can handle \textit{dynamic} scenes. 
We qualitatively test \model{}'s 4D reconstruction ability, showing examples of dynamic aligned pointmaps at multiple time steps in Figure \ref{fig:4d_reconstruction}. 
\model{} can be trained to achieve such results by finetuning a 16 static views checkpoint on the PointOdyssey~\cite{zheng2023pointodyssey} and TartanAir~\cite{tartanair2020iros} datasets, consisting of 110 dynamic and 150 static scenes, respectively. 
We freeze the ViT encoder, use 224x224 image resolution, and swap in a newly-initialized global DPT head. We fine-tune the model with 15 epochs with a frame length of 16, batch size per GPU of 1, and use the same learning rate schedule as \model{}. The process takes 45 hours to finetune on 2 Nvidia Quadro RTX A6000 GPUs.

We see that our approach produces qualitatively reasonable reconstructions with minimal changes.
MonST3R \cite{zhang2024monst3r} is a concurrent work also tackling dynamic scene reconstruction that builds atop DUSt3R. However, like DUSt3R, it assumes a pairwise architecture and also uses a separate model to predict optical flow. We show that the same \model{} architecture trained end-to-end with the same many-view pointmap regression (just swapping the data to dynamic scenes), can also work for 4D reconstruction. Importantly, our method remains significantly faster, opening the potential for real-time applications. 

\end{document}